\documentclass[10pt,twocolumn,letterpaper]{article}

\usepackage[pagenumbers]{wacv} 
\usepackage{multirow}
\usepackage{xcolor}
\usepackage{tikz}
\usepackage{amsmath}
\usepackage[table]{xcolor}
\usepackage{algorithm}
\usepackage{algpseudocode}
\usepackage{cuted}
\usepackage{adjustbox}

\definecolor{wacvblue}{rgb}{0.21,0.49,0.74}
\newcommand{\coloreddot}[1]{%
    \begin{tikzpicture}[baseline=-\the\dimexpr\fontdimen22\textfont2\relax]
        \filldraw[color=#1] (0,0) circle (2pt);
    \end{tikzpicture}%
}
\newcommand{\mypar}[1]{\noindent\textbf{#1}}

\definecolor{natural}{RGB}{180, 86, 87}
\definecolor{specialized}{RGB}{107, 163, 112}
\definecolor{structured}{RGB}{84, 114, 168}
\definecolor{lightorange}{rgb}{0.98, 0.95, 0.92}

\algnewcommand\algorithmicret{\textbf{Returns:}}
\algnewcommand\Returns{\item[\algorithmicret]}
\algnewcommand\algorithmicinit{\textbf{Initialization:}}
\algnewcommand\Init{\item[\algorithmicinit]}
\algnewcommand\algorithmforward{\textbf{Function:}}
\algnewcommand\Forward{\item[\algorithmforward]}
\algnewcommand\algorithmicinput{\textbf{Input:}}
\algnewcommand\Input{\item[\algorithmicinput]}

\usepackage[pagebackref,breaklinks,colorlinks,allcolors=wacvblue]{hyperref}
\usepackage{caption}

\title{PVeRA: Probabilistic Vector-Based Random Matrix Adaptation}

\author{
Leo Fillioux\textsuperscript{1,2} \and 
Enzo Ferrante\textsuperscript{3} \and 
Paul-Henry Cournède\textsuperscript{1,2} \and
Maria Vakalopoulou\textsuperscript{1,2} \and 
Stergios Christodoulidis\textsuperscript{1,2} 
\\
\\
\textsuperscript{1} MICS Laboratory, CentraleSupélec, Université Paris-Saclay \\
\textsuperscript{2} IHU PRISM, National Center for Precision Medicine in Oncology, Gustave Roussy \\
\textsuperscript{3} Institute of Computer Sciences, CONICET, Universidad de Buenos Aires
}

\begin{document}
\maketitle

\setlength{\stripsep}{0pt}
\begin{strip}
    \centering
    \vspace{-0.2cm}
    \href{https://github.com/leofillioux/pvera}{
        \adjustbox{valign=c}{
          \includegraphics[height=1.7\baselineskip]{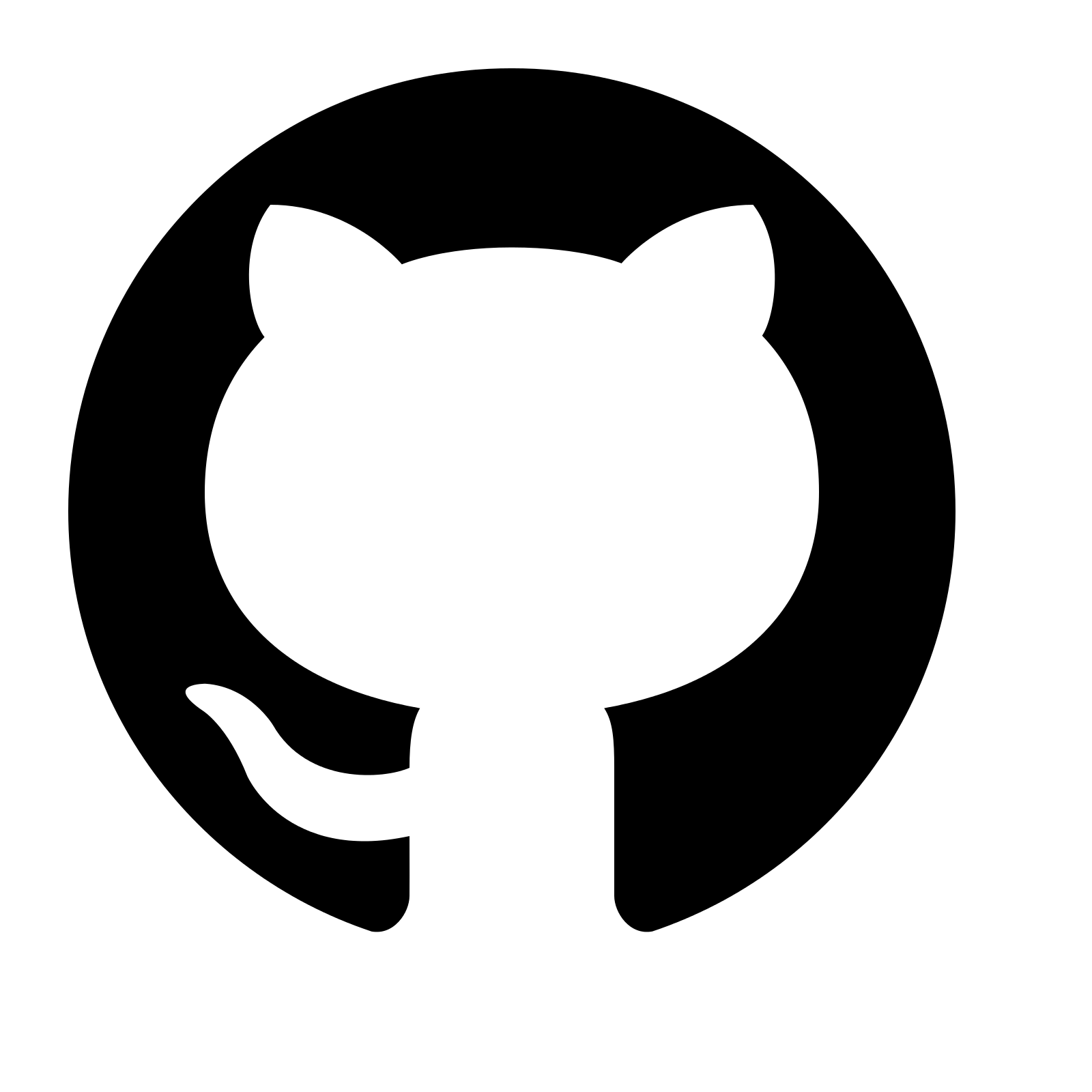}
        }
        {\large GitHub}
    }
    \hskip 0.2in
    \href{https://huggingface.co/docs/peft/main/package_reference/pvera}{
        \adjustbox{valign=c}{
          \includegraphics[height=2.2\baselineskip]{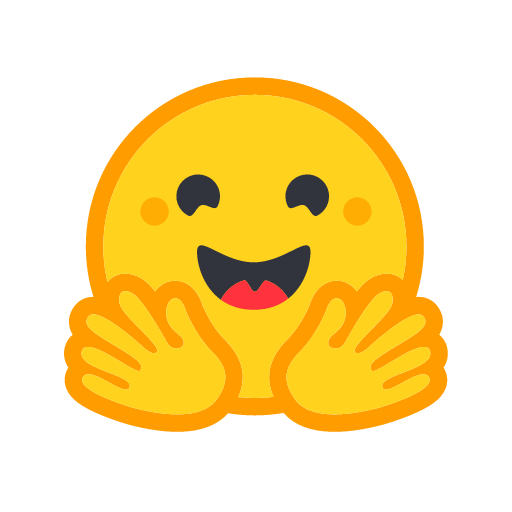}
        }
        {\large Hugging Face}
    }
    \vspace{0.4cm}
\end{strip}

\begin{abstract}
Large foundation models have emerged in the last years and are pushing performance boundaries for a variety of tasks. Training or even finetuning such models demands vast datasets and computational resources, which are often scarce and costly. Adaptation methods provide a computationally efficient solution to address these limitations by allowing such models to be finetuned on small amounts of data and computing power. This is achieved by appending new trainable modules to frozen backbones with only a fraction of the trainable parameters and fitting only these modules on novel tasks. Recently, the VeRA adapter was shown to excel in parameter-efficient adaptations by utilizing a pair of frozen random low-rank matrices shared across all layers. In this paper, we propose PVeRA, a probabilistic version of the VeRA adapter, which modifies the low-rank matrices of VeRA in a probabilistic manner. This modification naturally allows handling inherent ambiguities in the input and allows for different sampling configurations during training and testing. A comprehensive evaluation was performed on the VTAB-1k benchmark and seven adapters, with PVeRA outperforming VeRA and other adapters. Our code for training models with PVeRA and benchmarking all adapters is available \href{https://github.com/leofillioux/pvera}{here}.
\end{abstract}
\section{Introduction}

\begin{figure}
    \centering
    \begin{subfigure}{\linewidth}
        \centering
        \includegraphics[width=\textwidth]{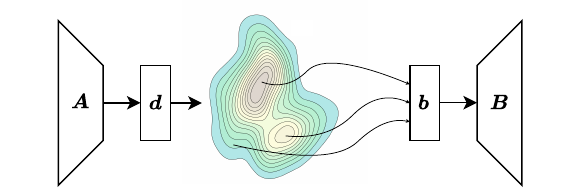}
        \caption{}
    \end{subfigure}
    \begin{subfigure}{\linewidth}
        \centering
        \includegraphics[width=\textwidth]{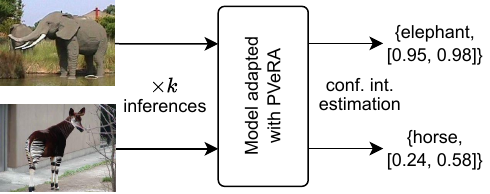}
        \caption{}
    \end{subfigure}
    \caption{\textbf{Probabilistic Vector-Based Random Matrix Adaptation}. (a) PVeRA learns a distribution of latent adaptations, from which samples are drawn to compute the adaptation. (b) We showcase how a model adapted with PVeRA can be used to estimate confidence intervals for the prediction.}
    \label{fig:teaser_figure}
\end{figure}

Large foundation models trained on vast datasets have appeared in the last years, pushing the performance boundaries for multiple tasks to unprecedented levels. Their exceptional performance is grounded on advances in self-supervised learning (SSL) and, as such, allows them to utilize rich and large-scale diverse datasets for their training. These models, characterized by their versatility across an impressive number of tasks, can be employed for a variety of downstream tasks even far from their training distribution. Additionally, due to their ability to learn meaningful general representations using self-supervision \cite{oquab2023dinov2, chen2020simple}, their performance can be satisfactory for a number of tasks also in zero-shot scenarios \cite{radford2021learning, kirillov2023segany}. Fine-tuning these models on specific datasets can further increase their performances, especially when there is a large domain shift between the original training data and the target data (e.g., medical tasks). Such fine-tuning, however, comes at the cost of computational requirements due to these models' large amount of trainable parameters while introducing an increased risk of model overfitting, especially in low data regimes.

To address these challenges, adapters have emerged as a lightweight alternative to traditional fine-tuning \cite{houlsby2019parameter}. Adapters, a type of parameter-efficient finetuning (PEFT), are modules with only a small number of trainable parameters that can be appended to large frozen pre-trained models, aiming at modifying the intermediate representations of such models towards improving the performance on a desired dataset. The training of such modules is performed by minimizing the objective function of a target task and by freezing the main part of the foundation model. The intuition is to utilize the general-purpose representations of large foundation models, which capture meaningful information across domains. Adapters, as plug-and-play modules, then act to bridge the domain gap between the original training data and the target data. An alternative to using adapters is to simply perform linear probing, which does not modify the intrinsic representations of the model, but which comes as a simple and extremely computationally efficient alternative (features can be precomputed). However, this does not support large domain shift (see Section~\ref{sec:experiments}).

A number of adapters have been proposed, ranging from the original bottleneck adapter \cite{houlsby2019parameter}, to IA$^3$ \cite{liu2022ia3}, LoRA \cite{hu2022lora}, and VeRA \cite{kopiczko2024vera} which all have a distinct approach to adapting large pretrained models. Although originally introduced for natural language processing, adapters are now commonly used for vision or vision-language models. We base our work on LoRA's insight into the low intrinsic rank of weight changes during model adaptation, combined with VeRA's approach to leveraging frozen random matrices for efficient adaptation. We hypothesize that a probabilistic formulation of the low-rank adaptation introduces an inductive bias to the model, assisting it in handling ambiguities in the feature spaces and hence allowing for non-deterministic adaptations during training.

\mypar{Contributions.} We (\textit{i}) propose PVeRA (\textbf{P}robabilistic \textbf{Ve}ctor-based \textbf{R}andom matrix \textbf{A}daptation), a parameter efficient adapter that learns a distribution over weight adaptations by approaching the low-rank decomposition through frozen random matrices in a probabilistic manner. We show that such a modification (\textit{ii}) outperforms the original VeRA while naturally extending its utility, allowing it to (\textit{iii}) quantify uncertainty (\textit{iv}) and keep its well-calibrated capabilities. (\textit{v}) We provide our code and implementation for easy plug-and-play adaptation with PVeRA and other adapters, allowing to reproduce training results. To support these claims, we conduct extensive experiments on the VTAB-1k benchmark, consisting of 19 datasets across three categories.
\section{Related Work}

\subsection{Adapters} 
Fine-tuning has emerged as a natural approach for transferring high-level features trained on larger datasets to other downstream tasks, based on the observation that learned features at the beginning of the network were more general \cite{yosinski2014transferable}. As the size of datasets and the available computational resources have increased, so has the size of deep neural networks. This results in increasingly impressive performance but also in more computationally demanding fine-tuning.

Adapters were originally introduced in natural language processing with the bottleneck adapter \cite{houlsby2019parameter} where a two-layer autoencoder is introduced at two locations in each Transformer encoder layer. Both autoencoders, together with normalization layers \cite{ba2016layer}, are trained. The Compacter adapter \cite{mahabadi2021compacter} builds upon the bottleneck adapter, adding parametrized hypercomplex multiplication layers to lower the number of parameters. Both adapters provide an interesting approach, helping to account for the domain shift in the data. However, they are directly added between layers of the main branch of the models, resulting in direct modification of the representations. Other adapters focus specifically on vision-language models \cite{elizalde2022clap, zhou2022learning, yao2023visual}. The AdaptFormer adapter \cite{chen2022adaptformer} takes a similar approach as the bottleneck adapter, and is placed only in one location of each Transformer encoder layer. Closer to our work, LoRA (Low-Rank Adaptation) \cite{hu2022lora} is an adapter that aims to mitigate the domain shift problem by approximating the change of weight in the query and value branches of the attention mechanism with low-rank components. The adapter computes an update of the frozen weights by a low-rank approximation $\Delta \boldsymbol W$. In contrast to the bottleneck adapters, LoRA is added in parallel to the main branch and only affects the representations by addition while it is also initialized accordingly such that the initial adaptation is zero, allowing for unchanged layers. Based on LoRA's work, VeRA (Vector-based Random matrix Adaptation) \cite{kopiczko2024vera} uses a pair of frozen low-rank decomposition matrices which are shared across all Transformer encoder layers and only trains scaling vectors, therefore drastically lowering the number of trained parameters. The (IA)$^3$~\cite{liu2022ia3} adapter takes a very different approach to other adapters, and learns three vectors per encoder layer, $l_v$ and $l_k$ which are applied (via element-wise multiplication) to the value and key components of the multi-head attention, as well as $l_{ff}$ which is used to adapt the last linear layer of the encoder layer. Like VeRA, this results in a very small number of trainable parameters. Recently, new methods have introduced sparse matrices to reduce the memory consumption of adapters and training time \cite{He2025SMT,sparse2024bhardwaj}.

Prompt tuning can also be considered as an adapter since it comes as a low-parameter alternative to fine-tuning. It consists of prepending a set of trainable prompts to the input of a Tranformer model or at each Transformer encoder layer. These input-independent tokens interact with the input tokens in the self-attention part of the Transformer encoder and will, therefore, participate in adapting the intermediate representations. Originally introduced in the context of natural language processing \cite{lester2021power}, it has also been extended to computer vision \cite{jia2022visual}. Prompt tuning can be seen as introducing an offset/bias in the latent representations in order to shift to a distribution closer to the training data. However, it has recently been proven to show limited expressiveness \cite{wang2024universality}.

\subsection{Probabilistic Deep Learning}
Probabilistic deep learning is a subfield that combines deep learning with probabilistic modeling, allowing models to learn stochastic functions rather than deterministic ones. Such formulations introduce an inductive bias, allowing the models to naturally capture ambiguities present in the inputs. One of the canonical models in probabilistic deep learning is the variational autoencoder (VAE) \cite{kingma2014vae}. VAEs are generative models in which the bottleneck of the autoencoder predicts a multivariate normal distribution, parameterized by a vector of means and standard deviations. Latent vectors are then sampled from this distribution. A KL divergence loss is employed to enforce a zero-mean and unit-variance prior on the learned normal distribution. Bayesian neural networks (BNNs) \cite{radford1996bayesian, mackay1992bnn} are another example of probabilistic deep learning models. Instead of learning fixed weights like standard neural networks, they learn a probability distribution over the weights. Training BNNs comes at a heavy computational cost, but they can be approximated using a standard neural network with dropout \cite{gal2016dropout}. The related concept of probabilistic embeddings refers to the formulations that map inputs to probability distributions within the embedding space, rather than simple point estimates \cite{oh2018modeling,chun2021probabilistic,shi2019probabilistic}. To the best of our knowledge, probabilistic formulations for adapters have not been extensively explored. ProbVLM \cite{upadhyay2023probvlm} is a post-host probabilistic adapter that is placed at the last layer of a pretrained model and expands the point estimates to probabilistic embeddings. Their proposed adapter estimates a probability distribution for the embeddings of a pretrained vision-language model while trying to remain faithful to the original embedding. However, unlike the versatility of the proposed method, this adapter is only applied at the very last layer of a single VLM model, while by default it cannot be applied on single modalities but only on multimodal tasks with both images and text available.
\section{Methods}
\label{sec:methods}

\subsection{Preliminaries}
The attention mechanism \cite{bahdanau2016neural} was first introduced in the context of natural language processing for neural translation before being used as a key component of Transformers \cite{ashish2017attention}. It serves as a way to dynamically capture the relative importance of different components in the input with respect to the output. Let $\boldsymbol x\in\mathbb R^{l\times d}$ be the input to the attention mechanism with $l\in \mathbb N^+_*$ and $d\in\mathbb N^+_*$ being the sequence length and the dimensionality of the feature space respectively. Self-attention is a particular case of the attention mechanism where the inputs to the query, value, and key branches come from the same input $\boldsymbol x$ and are obtained using a linear layer with weights $\boldsymbol W_q$, $\boldsymbol W_k$, $\boldsymbol W_v\in \mathbb R^{d\times d}$ and biases $\boldsymbol b_{W_q}$, $\boldsymbol b_{W_k}$, $\boldsymbol b_{W_v} \in \mathbb R^d$ respectively. The output of the self-attention is then obtained by a scaled dot product of the query, key, and value components , referred to in this section as $q$, $k$, and $v$, respectively. Note we will use $\boldsymbol x_{\{q, k, v\}}$ to refer to the $q$, $k$, and $v$ components of $\boldsymbol x$ (and other elements) in a more compact form.

\begin{equation}
\boldsymbol x_{\{q, k, v\}} = \boldsymbol x \boldsymbol W^T_{\{q, k, v\}} + \boldsymbol b_{W_{\{q, k, v\}}}
\end{equation}

\begin{equation}
\text{Attention}(\boldsymbol x_q, \boldsymbol x_k, \boldsymbol x_v) = \text{softmax}\Big(\frac{\boldsymbol x_q \boldsymbol x_k^T}{\sqrt{d}}\Big)\boldsymbol x_v
\end{equation}

LoRA \cite{hu2022lora} is applied to the query and value branches of the self-attention mechanism. In more detail, given a rank $r\in\mathbb N^+_* < d$, and a fixed scaling parameter $\alpha\in \mathbb R^+_*$, LoRA's only trainable parameters are $\boldsymbol A \in \mathbb R^{d\times r}$ and $\boldsymbol B \in \mathbb R^{r\times d}$, which constitute a two-layer downsample-upsample adapter in parallel to the main branch.

\begin{multline}
\label{eq:lora}
\boldsymbol x_{\text{LoRA}\{q, v\}} =
\big(\boldsymbol x \boldsymbol W^T_{\{q, v\}} + \boldsymbol b_{W_{\{q, v\}}}\big)\\
+
\frac{\alpha}{r}
\big( \boldsymbol x \boldsymbol A_{\{q, v\}} \boldsymbol B_{\{q, v\}} \big)
\end{multline}

\begin{figure}[t]
  \centering
  \includegraphics[width=\linewidth]{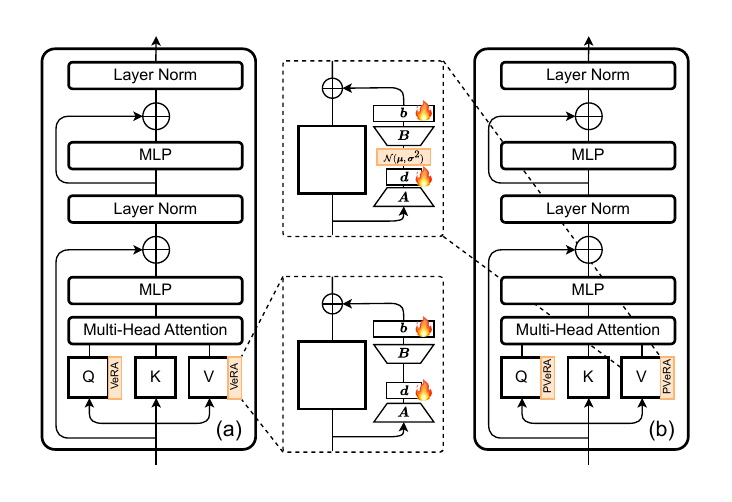}
  \caption{\textbf{Representation of the VeRA and PVeRA architectures}. (a) VeRA~\cite{kopiczko2024vera} on one Transformer encoder layer. (b) Our proposed PVeRA: a probabilistic variation of VeRA applied to the query and value components on the multi-head attention mechanism of the Transformer encoder layer. Pseudocode for PVeRA is shown in Appendix Section \ref{supp_sec:pseudocode}.}
\label{fig:proposed_architecture}
\end{figure}

As shown in Figure \ref{fig:proposed_architecture}, VeRA \cite{kopiczko2024vera} is also applied both to the query and value branches. The $\boldsymbol A$ and $\boldsymbol B$ matrices are both initialized to $\mathcal N(\boldsymbol 0, \boldsymbol \sigma^2)$ and shared across all layers of the Transformer encoder (therefore allowing to greatly lower the number of parameters of the adapter), with a precomputed $\boldsymbol \sigma$. The only trainable parameters are $\boldsymbol d \in \mathbb R^r$ and $\boldsymbol b \in \mathbb R^d$.

\begin{multline}
\boldsymbol x_{\text{VeRA}\{q, v\}} =
\big(\boldsymbol x \boldsymbol W^T_{\{q, v\}} + \boldsymbol b_{W_{\{q, v\}}}\big) \\
+
\alpha
\Big( \big( \boldsymbol x \boldsymbol A_{\{q, v\}} \odot \boldsymbol d_{\{q, v\}} \big) \boldsymbol B_{\{q, v\}} \odot \boldsymbol b_{\{q, v\}} \Big)
\end{multline}

\noindent
Note that LoRA and VeRA have different approaches to the scaling parameter. LoRA uses $\frac{\alpha}{r}$ to adjust to different learning rates, while VeRA uses $\alpha$ but uses a different learning rate for the adapter and the prediction head.

\subsection{Probabilistic Adaptation}
Our proposed adapter, PVeRA, is a probabilistic adaptation of VeRA. $\boldsymbol A_{\{q, v\}} \in \mathbb R^{d\times 2r}$ and $\boldsymbol d_{\{q, v\}} \in \mathbb R^{2r}$ are used to generate $\boldsymbol \mu_{\{q, v\}} \in \mathbb R^r$ and $\boldsymbol \sigma_{\{q, v\}} \in \mathbb R^r$, representing the mean and standard deviation of a multivariate normal distribution, respectively. Using the reparameterization trick, we sample from the learned distribution of the latent space $\boldsymbol z_{\{q, v\}}\sim\mathcal N(\boldsymbol \mu_{\{q, v\}}, \boldsymbol \sigma_{\{q, v\}}^2)$ as the input to $\boldsymbol B_{\{q, v\}} \in \mathbb R^{d\times r}$ and $\boldsymbol b \in \mathbb R^d$.

\begin{equation}
    \boldsymbol \mu_{\{q, v\}}, \boldsymbol \sigma_{\{q, v\}} = \boldsymbol x \boldsymbol A_{\{q, v\}} \odot \boldsymbol d_{\{q, v\}}
\end{equation}

\begin{equation}
    \boldsymbol z_{\{q, v\}} = \boldsymbol \epsilon \odot \boldsymbol \sigma_{\{q, v\}} + \boldsymbol \mu_{\{q, v\}} \text{, with } \boldsymbol \epsilon \sim \mathcal N(\boldsymbol 0, \boldsymbol 1)
\end{equation}

\begin{multline}
\boldsymbol x_{\text{PVeRA}\{q, v\}} =
\big(\boldsymbol x \boldsymbol W^T_{\{q, v\}} + \boldsymbol b_{W_{\{q, v\}}}\big)
\\ +
\alpha (
\boldsymbol z_{\{q, v\}} \boldsymbol B_{\{q, v\}} \odot \boldsymbol b_{\{q, v\}})
\end{multline}

We use a Kullback-Leibler divergence loss to enforce a standard Normal prior to each PVeRA adapter. With $\beta \in \mathbb R_*^+$ defined as the KL loss scaling factor (defined per dataset using a grid search on the validation loss), we define the total loss as follows.

\begin{equation}
    \mathcal{L_\text{total}} = \mathcal{L_\text{classification}} + \beta \sum_{\text{layer}\in\text{ViT}} \mathcal{L}_{\text{KL, layer}}
\end{equation}

\begin{equation}
\mathcal{L}_{\text{KL, layer}} = \frac{1}{2} \sum_{i\in\{q, v\}}D_\text{KL} \big(\mathcal N(\boldsymbol \mu_\text{i, layer}, \boldsymbol \sigma^2_\text{i, layer})~\Vert~\mathcal N(\boldsymbol 0 , \boldsymbol I)\big)
\end{equation}

\subsection{Inference}
During inference, the adaptation can either be performed in a deterministic or probabilistic fashion. Firstly, a deterministic sampling can be considered using $\boldsymbol z_{\{q, v\}} = \boldsymbol \mu_{\{q, v\}}$ instead of $\boldsymbol z_{\{q, v\}} = \boldsymbol \epsilon \odot \boldsymbol \sigma_{\{q, v\}} + \boldsymbol \mu_{\{q, v\}}$. As such, the weights can be easily merged into the original weights of the model, resulting in no additional inference time, as for VeRA. This can be achieved by taking $\boldsymbol A_\mu$, the half of ($\boldsymbol A \odot \boldsymbol d$) responsible for generating $\boldsymbol \mu$ and assigning the new weight to the linear layer.

\begin{equation}
    \boldsymbol W_{\{q, v\}} \leftarrow \boldsymbol W_{\{q, v\}} + \alpha \boldsymbol A_{\mu\{q, v\}} \boldsymbol B_{\{q, v\}} \odot \boldsymbol b_{\{q, v\}}
\end{equation}
We use this sampling strategy in our experiments. Alternatively, a probabilistic adaptation can be considered such that the adaptations are randomly drawn from the learned distribution. Such a mode can be used in the context of Monte Carlo confidence interval estimation. Some results on inference time sampling are shown in Section \ref{sec:experiments}. During training, we sample from the distribution, but during inference both methods are possible. Unless otherwise stated, we use the deterministic inference during our experiments.
\section{Experiments}
\label{sec:experiments}
We evaluate PVeRA on the VTAB-1k benchmark~\cite{zhai2020vtab1k}, and compare the performance with seven other adapters. We use DINOv2~\cite{oquab2023dinov2} (ViT-B/14) as a backbone and freeze every layer except for the adapter and the linear probe. DINOv2 is a framework for training models on images. We decide to adapt DINOv2 rather than a ViT trained on ImageNet, as these SSL foundation models have become the state-of-the-art and are used in most downstream applications. In Appendix Section~\ref{supp_sec:nlp}, we additionally provide a benchmark for PVeRA on a few language tasks to further highlight its utility and superiority.

\subsection{Baseline Adapters}
We compare PVeRA with six other adapters and linear probing. Here is a brief description of these adapters.

\mypar{Linear probing} consists of a linear layer (mapping from the dimensionality of the feature space to the number of classes of the dataset) and a softmax activation function. The same classification head is used across all experiments.

\mypar{Bottleneck}~\cite{houlsby2019parameter} was the first introduced adapter; it consists of a two-layer downsample-upsample architecture, which is introduced at two locations in each Transformer encoder layer after each MLP. We use a ReLU activation function and perform a grid search across the reduction ratios.

\mypar{AdaptFormer}~\cite{chen2022adaptformer} has a similar structure to the bottleneck adapter, except that only one such adapter is introduced in each layer: in parallel to the LayerNorm and the MLP. We use a ReLU activation function and perform a grid search across the reduction ratios.

\mypar{(IA)$^3$}~\cite{liu2022ia3} learns scaling vectors at three locations of the encoder layer: for the key and value elements of the attention mechanism, and for the MLP. It does not have any hyperparameters, but we perform grid search on the learning rate.

\mypar{DoRA}~\cite{liu2024dora} is based on LoRA, but decomposes the pretrained weights into magnitude and direction and instead learns these components. We use a scaling parameter of $\alpha=16$, and perform a grid search across the ranks.

\mypar{LoRA}~\cite{hu2022lora} has been described in Section~\ref{sec:methods}. We use a scaling value of $\alpha=16$ and make a grid search across the ranks.

\mypar{VeRA}~\cite{kopiczko2024vera} has been described in Section~\ref{sec:methods}. We use a scaling value of $\alpha=16$. As in the original paper, we do not scale $\alpha$ with the rank, but instead fix the rank to a rank larger than LoRA ($r=256$), and instead perform a grid search on the adapter-specific learning rate. The linear probe learning rate is the same as for the other adapters.

\mypar{PVeRA} was introduced in Section~\ref{sec:methods}. We use the same approach as the one described for VeRA.

We selected linear probing, Bottleneck, (IA)$^3$, and AdaptFormer as they represent widely used baseline adapters. Our main comparison, however, focuses on LoRA, DoRA, and VeRA, low-rank adaptation techniques derived from LoRA, as PVeRA builds upon this family of methods.

\subsection{Experimental Setup}

\mypar{VTAB-1k benchmark.} The VTAB-1k~\cite{zhai2020vtab1k} benchmark consists of 19 datasets~\cite{li2006one, cifar10, cimpoi14describing, Nilsback08, parkhi12a, xiao2010sun, netzer2011reading, veeling2018rotation, helber2017eurosat, cheng2017remote, kaggle-diabetic-retinopathy, johnson2017clevr, beattie2016deepmind, dsprites17, Geiger2013IJRR, lecun2004learning} to evaluate PEFT methods in a few-shot scenario for classification. In each dataset, the training set only consists of 1000 labeled samples (800 for training and 200 for validation), which evaluates the capability of PEFT methods to learn in a low-data regime. The size of the test set depends on the dataset. The datasets come from varied sources, and are divided in three categories: natural, specialized, and structured.

\mypar{Grid search.} We perform a grid search across all trainings (i.e., different seeds could have different values of optimal grid search value). We limit our grid search to the main hyperparameter (reduction ratio, rank, or learning rate), and to three values. See Appendix Section \ref{supp_sec:grid_search} for more detail on the grid search.

\mypar{Implementation details.}
All trainings are performed over three random seeds, and the resulting test accuracies are averaged. We train with a batch size of $16$, for $500$ epochs with early stopping with a patience of $20$ epochs, and a relative tolerance of $\epsilon=0.1\%$. Models are trained with the AdamW~\cite{loshchilov2019decoupledweightdecayregularization} optimizer, and cross entropy as a loss, with a learning rate of $10^{-4}$ (except for linear probing, (IA)$^3$, and for VeRA and PVeRA for which the linear probe and the adapters have different learning rates, similarly to the original VeRA paper) and a weight decay of $10^{-4}$.

\subsection{Results}

\mypar{Comparison to the baselines.}
In Table~\ref{tab:main}, we show the results of PVeRA against linear probing and six other adapters. While some adapters show strong performance across some datasets (e.g., DoRA scoring best performance across seven datasets), PVeRA shows the most stable performance with the highest average performance, while keeping a very low number of trainable parameters. It is worth noting that due to the high generalizability of DINOv2, the performance on the natural images (datasets marked with \coloreddot{natural} in Table~\ref{tab:main}) is quite high, sometimes even higher than some adapters, especially those not operating on the attention mechanism but in a more direct manner (Bottleneck and AdaptFormer). However, the linear probing performance of DINOv2 is dropped when there is a larger distribution shift (datasets marked with \coloreddot{specialized} and \coloreddot{structured} in Table~\ref{tab:main}). See Appendix Section \ref{supp_sec:detailed_results} for a more detailed table showing statistical tests and standard deviations, as well as similar results for a smaller model (DINOv2 ViT-S/14) and a larger model (DINOv2 ViT-L/14).
\begin{table*}[]
    \centering
    \resizebox{\textwidth}{!}{
    \begin{tabular}{l c c c c c c c c c c c c c c c c c c c c}
    \toprule
    
         & \rotatebox[origin=l]{90}{\coloreddot{natural}~Caltech101}
         & \rotatebox[origin=l]{90}{\coloreddot{natural}~CIFAR-100}
         & \rotatebox[origin=l]{90}{\coloreddot{natural}~DTD}
         & \rotatebox[origin=l]{90}{\coloreddot{natural}~Flowers102}
         & \rotatebox[origin=l]{90}{\coloreddot{natural}~Pets}
         & \rotatebox[origin=l]{90}{\coloreddot{natural}~Sun397}
         & \rotatebox[origin=l]{90}{\coloreddot{natural}~SVHN}
         
         & \rotatebox[origin=l]{90}{\coloreddot{specialized}~Camelyon}
         & \rotatebox[origin=l]{90}{\coloreddot{specialized}~EuroSAT}
         & \rotatebox[origin=l]{90}{\coloreddot{specialized}~Resisc45}
         & \rotatebox[origin=l]{90}{\coloreddot{specialized}~Retinopathy}

         & \rotatebox[origin=l]{90}{\coloreddot{structured}~Clevr-Count}
         & \rotatebox[origin=l]{90}{\coloreddot{structured}~Clevr-Dist}
         & \rotatebox[origin=l]{90}{\coloreddot{structured}~DMLab}
         & \rotatebox[origin=l]{90}{\coloreddot{structured}~dSpr-Loc}
         & \rotatebox[origin=l]{90}{\coloreddot{structured}~dSpr-Ori}
         & \rotatebox[origin=l]{90}{\coloreddot{structured}~KITTI-Dist}
         & \rotatebox[origin=l]{90}{\coloreddot{structured}~sNORB-Azim}
         & \rotatebox[origin=l]{90}{\coloreddot{structured}~sNORB-Elev}

         & \rotatebox[origin=l]{90}{\coloreddot{white}~Average}
         \\
    \midrule
         Linear & 85.1
                & 61.6
                & 73.7
                & \textbf{99.7}
                & 94.0
                & 51.3
                & 38.9
                & 82.3
                & 90.6
                & 78.6
                & 73.8
                & 42.9
                & 33.7
                & 41.2
                & 11.8
                & 31.0
                & 54.0
                & 12.4
                & 25.5
                & 57.0 \\
         Bottleneck & 88.6
                    & 41.8
                    & 72.3
                    & 98.7
                    & 85.8
                    & 32.5
                    & \underline{90.9}
                    & 86.4
                    & 94.3
                    & 82.7
                    & 73.6
                    & \underline{78.6}
                    & 60.5
                    & 48.1
                    & \underline{80.4}
                    & 49.9
                    & 79.7
                    & 19.1
                    & 31.7
                    & 68.2 \\
         (IA)$^3$ & 86.8
                & \underline{71.3}
                & \textbf{77.2}
                & \textbf{99.7}
                & \textbf{94.1}
                & 54.2
                & 68.7
                & 83.2
                & 93.0
                & 84.9
                & 74.4
                & 55.9
                & 51.8
                & 43.3
                & 55.5
                & \textbf{53.1}
                & 77.7
                & 14.4
                & 26.8
                & 66.6 \\
         AdaptFormer & 87.9
                     & 56.8
                     & 72.9
                     & 98.5
                     & 90.5
                     & 30.7
                     & 89.3
                     & \underline{86.5}
                     & 94.3
                     & 83.6
                     & 73.6
                     & \textbf{86.7}
                     & \textbf{61.7}
                     & 49.3
                     & 69.4
                     & 52.2
                     & 82.7
                     & 19.6
                     & \underline{36.7}
                     & 69.6 \\
         DoRA & \underline{90.0}
              & 61.8
              & 74.3
              & 99.1
              & 90.7
              & 49.1
              & \textbf{91.0}
              & \textbf{87.2}
              & \textbf{95.5}
              & \textbf{87.7}
              & 73.6
              & 63.7
              & 60.5
              & \textbf{52.6}
              & \textbf{84.3}
              & 52.1
              & 82.1
              & \textbf{21.2}
              & 32.2
              & \underline{71.0} \\
         LoRA & \textbf{90.1}
              & 64.5
              & 75.3
              & 99.5
              & 91.8
              & 50.7
              & 87.7
              & 84.7
              & \underline{94.9}
              & 84.9
              & \textbf{75.0}
              & 66.2
              & 59.5
              & \underline{50.3}
              & 76.8
              & \underline{52.3}
              & 80.8
              & 19.6
              & 34.3
              & 70.5 \\
         VeRA & 87.3
              & 70.1
              & \underline{76.6}
              & \textbf{99.7}
              & \textbf{94.1}
              & \underline{54.5}
              & 87.8
              & 84.3
              & 93.6
              & 84.9
              & \textbf{75.0}
              & 58.1
              & 58.6
              & 47.4
              & 73.8
              & 48.0
              & \textbf{84.0}
              & 19.0
              & 32.4
              & 69.9 \\
         \rowcolor{lightorange} PVeRA & 88.7
                                      & \textbf{71.7}
                                      & 76.1
                                      & \textbf{99.7}
                                      & 93.3
                                      & \textbf{54.7}
                                      & 88.7
                                      & 85.0
                                      & 94.3
                                      & \underline{86.4}
                                      & 74.8
                                      & 71.5
                                      & \underline{60.6}
                                      & 48.6
                                      & 72.1
                                      & 49.5
                                      & \underline{83.3}
                                      & \underline{20.3}
                                      & \textbf{37.0}
                                      & \textbf{71.4} \\
    \bottomrule
    \end{tabular}}
    \caption{\textbf{Benchmarking PVeRA against VeRA and other adapters on VTAB-1k}. We benchmark PVeRA against other adapters on the 19 datasets of VTAB-1k (seven natural datasets, four specialized datasets, and eight structured datasets). Reported results are the average accuracy (\%) across three random seeds. The best results are indicated in \textbf{bold}, and second best results are \underline{underlined}.}
    \label{tab:main}
\end{table*}

\begin{figure*}
    \centering
    \begin{subfigure}{0.23\textwidth}
        \centering
        \includegraphics[width=\textwidth]{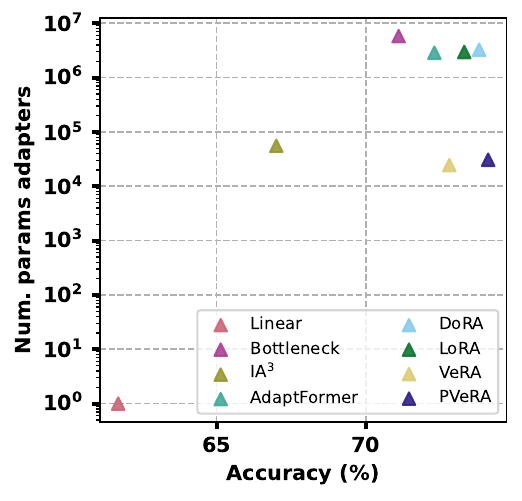}
        \caption{Num. parameters adapters}
    \end{subfigure}
    \begin{subfigure}{0.23\textwidth}
        \centering
        \includegraphics[width=\textwidth]{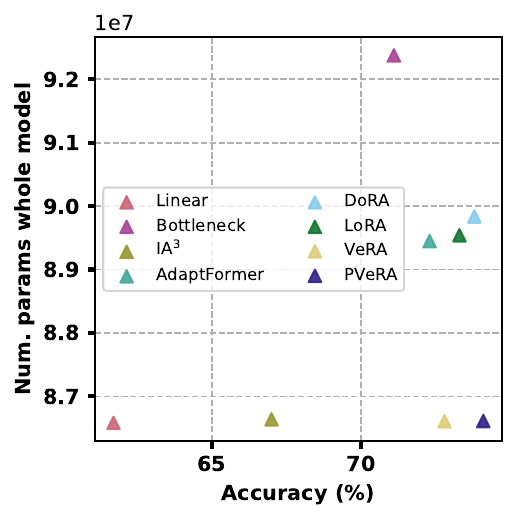}
        \caption{Num. parameters whole model}
    \end{subfigure}
    \begin{subfigure}{0.23\textwidth}
        \centering
        \includegraphics[width=\textwidth]{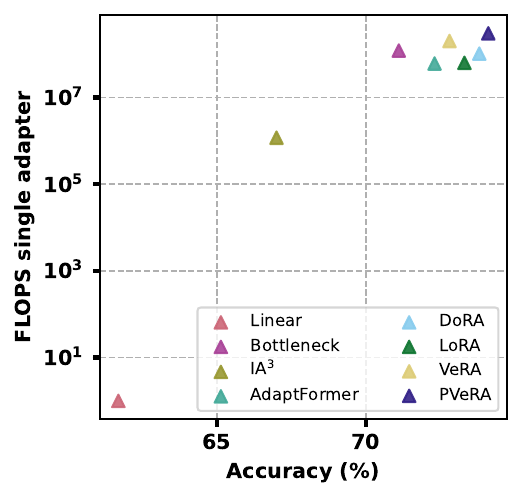}
        \caption{Single adapter FLOPS}
    \end{subfigure}
    \begin{subfigure}{0.23\textwidth}
        \centering
        \includegraphics[width=\textwidth]{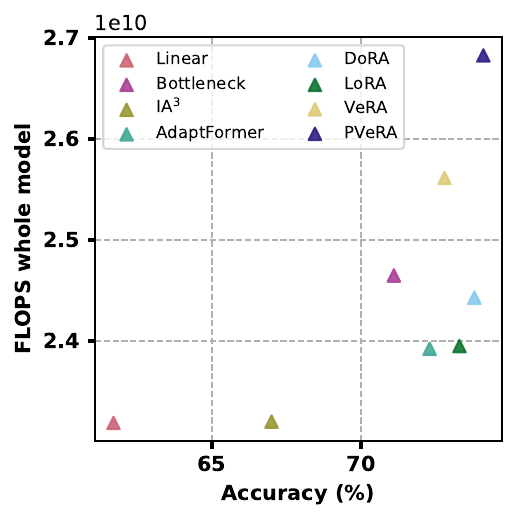}
        \caption{Whole model FLOPS}
    \end{subfigure}
    \caption{\textbf{Comparison of the computation efficiency}.
    (a) Number of trainable parameters of the adapters against the accuracy.
    (b) Number of parameters of the whole model adapted with each adapter against the accuracy.
    (c) FLOPS of a single adapter against the accuracy.
    (d) FLOPS of a whole model adapter with each adapter against the accuracy.
    Note that for the adapters for which a grid search over the hyperparameters is performed, the value of the number of parameters and FLOPS represents the average of the parameters and FLOPS respectively, weighted by the proportion of each chosen hyperparameter (see Appendix Section~\ref{supp_sec:grid_search}).}
    \label{fig:computation_efficiency}
\end{figure*}

\mypar{Computational efficiency.} While we show that PVeRA outperforms VeRA and other adapters on average, it is interesting to analyze the computational efficiency of the adapter. Figure~\ref{fig:computation_efficiency}.a and Figure~\ref{fig:computation_efficiency}.b plot the performance with the number of trainable parameters of adapters and number of parameters of the whole adapted model respectively. It is quite clear that PVeRA boosts the performance of VeRA, while retaining its low number of parameters, remaining orders of magnitude smaller than most other adapters. Figure~\ref{fig:computation_efficiency}.c and Figure~\ref{fig:computation_efficiency}.d compare the floating-point operations per second (FLOPS) for a single adapter and the whole adapted model respectively. While PVeRA does increase the FLOPS of the base model by $15.7\%$, compared to $10.5\%$ for VeRA, it is important to note that similarly to LoRA and VeRA, the weights of PVeRA can be incorporated into updated weight of the ViT for inference (see Section~\ref{sec:methods}), leading to no computational overhead. This is not the case for adapters such as Bottleneck or AdaptFormer, which maintain their computational overhead during inference.

\begin{figure}
    \centering
    \includegraphics[width=0.9\linewidth]{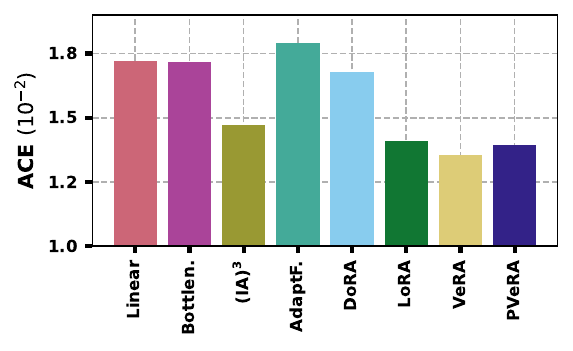}
    \caption{\textbf{Average calibration performance of adapters.} Average ACE across all datasets for all considered adapters. Lower is better.}
    \label{fig:calibration}
\end{figure}

\mypar{Calibration.}
Model calibration has gained a lot of attention in deep learning lately, taking root in earlier works applied to classical machine learning methods \cite{platt1999probabilistic, zadrozny2001obtaining}. The concept of calibration is for a model to not only predict accurately but also for the output scores to provide an accurate measure of confidence about the prediction. For a well-calibrated model, the output score can be interpreted as the actual probability of belonging to the positive class. Recent work has shown that although modern deep learning models show high accuracy, they are often poorly calibrated \cite{guo2017calibration}. We explore calibration because a method that improves performance should not come at the cost of calibration. The Expected Calibration Error (ECE) \cite{naeini2015obtaining} is a metric that is employed for quantifying the quality of the calibration, which takes the highest probability among the softmax output for each prediction on the test set (containing $N$ data points) and discretizes it into $B$ fixed-interval bins, with each bin $b$ containing $n_b$ data points. The ECE is defined as the mean difference between the accuracy and the mean prediction in each bin. For a perfectly calibrated model, the accuracy is equal to the mean prediction in each bin (called the confidence).
\begin{equation}
    \text{ECE} = \frac{1}{N} \sum_{b\in B} n_b \big\vert \text{accuracy}(b) - \text{confidence}(b) \big\vert
\end{equation}
In a multiclass classification setting, the lower probability bins will be empty as the lowest possible value for the highest probability is $\frac{1}{C}$ for a dataset with $C$ classes. The Adaptive Calibration Error (ACE) \cite{mukhoti2020calibrating} is based on the ECE, and spaces the bins for each one to contain an equal number of elements ($\forall b\in B, n_b=\frac{N}{B}$), therefore focusing less on the regions without any prediction. Figure~\ref{fig:calibration} compares the ACE across all adapters (geometric mean across all datasets). It shows that all LoRA-based methods (LoRA, VeRA, PVeRA) are better calibrated than other adapters. PVeRA improves the overall performance of VeRA, while keeping its capability to deliver well-calibrated models.

\begin{figure}
    \centering
    \begin{subfigure}{0.49\linewidth}
        \centering
        \includegraphics[width=\textwidth]{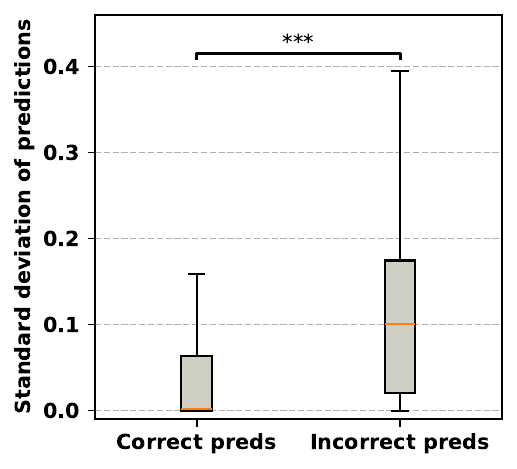}
        \caption{4 samples}
    \end{subfigure}
    \begin{subfigure}{0.49\linewidth}
        \centering
        \includegraphics[width=\textwidth]{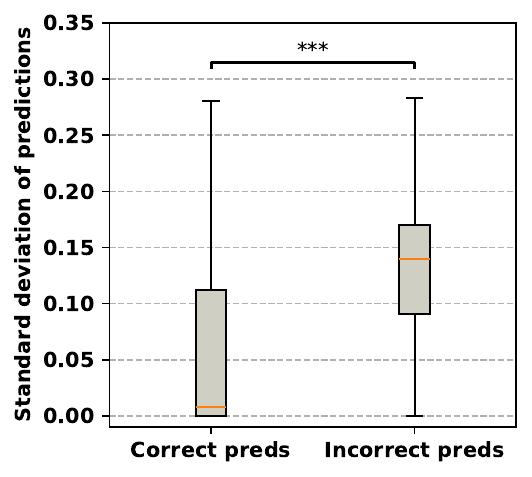}
        \caption{16 samples}
    \end{subfigure}
    \caption{\textbf{Uncertainty estimation visualization}. Distribution of standard deviation of the softmax scores for correctly and incorrectly classified samples when using (a) 4 samples and (b) 16 samples. Results across all datasets. The significance levels correspond to p-values for a one-sided unpaired Wilcoxon test, and indicate distributions with significantly different values.}
    \label{fig:uncertainty}
\end{figure}

\mypar{Uncertainty quantification.} The inference scheme used for all experiments is to sample from the learned distribution during training, and to use $\boldsymbol \mu_{\{q, v\}}$ during validation. The main intuition behind using a probabilistic adapter is to be able to inherently model the uncertainty of the model. One way to use the probabilistic nature of PVeRA to estimate uncertainty, is to perform multiple passes through the model while sampling from the learned distributions of the adapters. Multiple passes through the model will lead to different softmax scores. Looking at the distributions of these softmax scores gives an estimation of the uncertainty of the model. Algorithm~\ref{alg:uncertainty} explains the experimental setup, and Figure~\ref{fig:uncertainty} shows the difference in distribution. For wrong predictions there is a higher standard deviation in the predicted softmax, which suggests that the learned latent distributions capture uncertainty in the predictions. While this is naturally more computationally expensive, as multiple passes through the model are needed, it may be of use in more sensitive applications, for which robustness is more important than time efficiency.

\begin{algorithm}
\caption{\textbf{Uncertainty estimation algorithm}. Algorithm for estimating the uncertainty using a model adapted with PVeRA.}
\label{alg:uncertainty}
\begin{algorithmic}

\Input model $\mathcal M$, dataset $\mathcal D$, number of samples $k$

\Forward
\State sds $\gets$ []
\State accuracies $\gets$ []
\State $\mathcal M \gets$ enable\_inference\_sampling($\mathcal M$) 

\For{$x,y$ in $\mathcal D$}

\State pred $\gets$ []
\State acc $\gets$ []

\For{$i$ in $\{1,\dots,k\}$}
\State $\hat{y}\gets \mathcal M(x)$
\State acc.append(\texttt{argmax}($\hat{y}$) = $y$)
\State pred.append(\texttt{max}($\hat{y}$))
\EndFor

\State sds.append(\texttt{std}(pred))
\State accuracies.append(\texttt{mean}(acc))

\EndFor

\Returns
\State sd\_incorrect = stds[acc $>$ 0.5]
\State sd\_correct = stds[acc $\leq$ 0.5]

\end{algorithmic}
\end{algorithm}

\vspace{-1em}

\mypar{Confidence interval generation}. We can generate confidence intervals for the predictions, using the softmax scores from multiple passes, using the scores from the most predicted class, and estimating the confidence interval using the $t$-scores. We compared the width of the confidence intervals of softmax scores for correct and incorrect samples and found that correctly classified samples had much narrower confidence intervals than those incorrectly classified ($\boldsymbol{0.085}$ vs. $\boldsymbol{0.225}$). While there can be cases where the model is certain about incorrect predictions and uncertain about correct predictions, since the models perform overall well on the VTAB-1k benchmark, this is not the majority of cases.
In Appendix Section~\ref{supp_sec:uncertainty}, we further analyze generating Monte Carlo confidence intervals. Note that the interpretation of these confidence intervals as probability of belonging to the predicted class is conditioned on the calibration of the model.

\subsection{Feature Space Analysis}
\begin{figure}
    \centering
    \begin{subfigure}{0.49\linewidth}
        \centering
        \includegraphics[width=\textwidth]{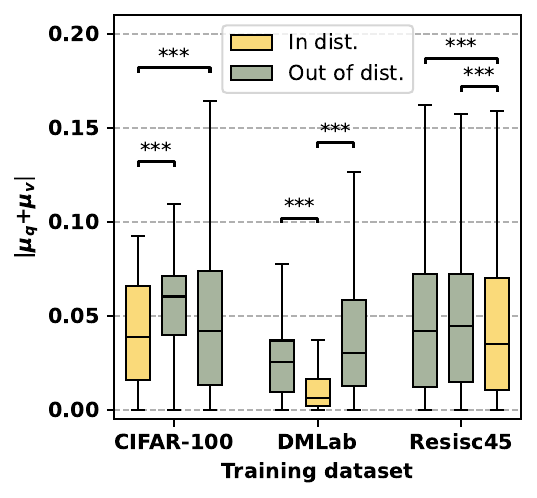}
        \caption{PVeRA}
    \end{subfigure}
    \begin{subfigure}{0.49\linewidth}
        \centering
        \includegraphics[width=\textwidth]{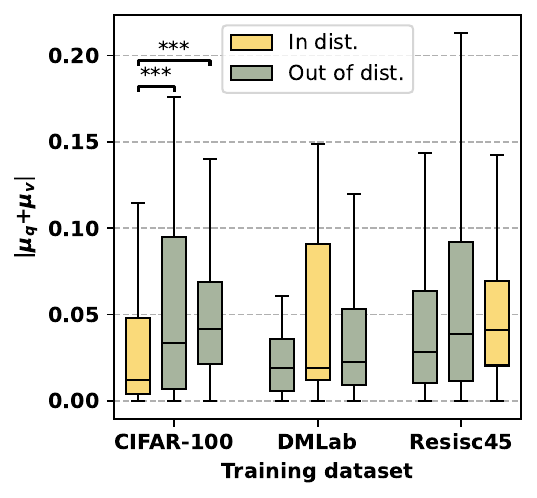}
        \caption{VeRA}
    \end{subfigure}
    \caption{\textbf{Out-of-distribution detection}. Distribution of $(\boldsymbol \mu_q + \boldsymbol \mu_v)$ for PVeRA (a) and VeRA (b) when testing learned models in distribution and out-of-distribution. The significance levels correspond to p-values for a one-sided unpaired Wilcoxon test, and indicate distributions with significantly lower values.}
    \label{fig:ood}
\end{figure}

\mypar{Out-of-distribution detection.} PVeRA adapters learn a distribution of adaptations instead of learning a deterministic adaptation. We can use these distributions for out-of-distribution detections instead of using Monte Carlo sampling. In Figure~\ref{fig:ood}, we analyze how the values of $(\boldsymbol \mu_q + \boldsymbol \mu_v)$ in the last layer of the adapter (as it is the closest to the prediction head) can be used for out-of-distribution detection. We take three models trained on three datasets, one from each dataset category. We test these models on each of these three datasets, leading to one in distribution and two out-of-distribution results. For PVeRA (Figure~\ref{fig:ood}.a), we find that the values $(\boldsymbol \mu_q + \boldsymbol \mu_v)$ of adaptation are significantly lower for in distribution than out-of-distribution, which is not the case for VeRA (Figure~\ref{fig:ood}.b). This indicates that these values can be used for detecting out-of distribution datasets. We hypothesize that adjusting the enforced prior of the PVeRA adapter could enhance this out-of-distribution detection.

\begin{figure}
    \centering
    \includegraphics[width=\linewidth]{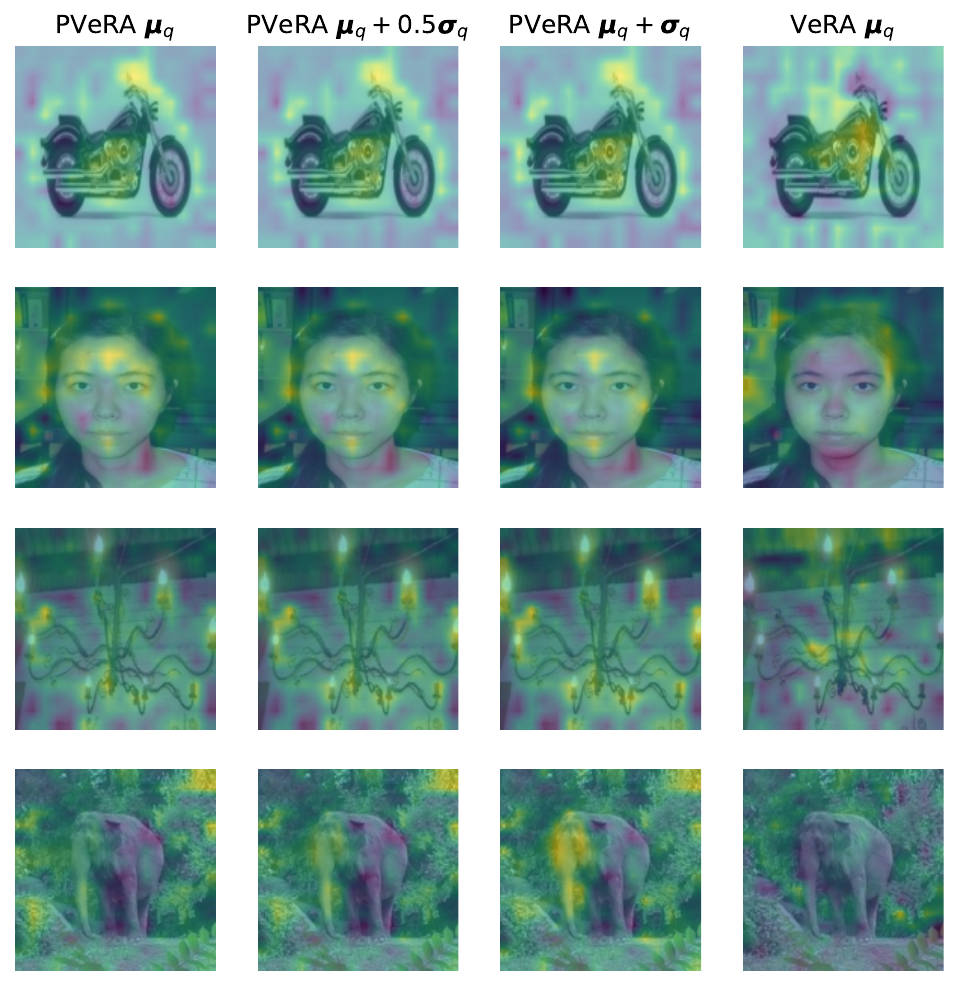}
    \caption{\textbf{Latent space projection for VeRA and PVeRA on Caltech101}. Five samples from the Caltech101 dataset, along with the projection of $\boldsymbol \mu_q$ for VeRA, and different draws from the $\mathcal N (\boldsymbol \mu_q, \boldsymbol \sigma^2_q)$ distribution for PVeRA (all from the last layer of the ViT).}
    \label{fig:latent_space}
\end{figure}

\mypar{Latent space projection.} In Figure~\ref{fig:latent_space} we visualize the adapters of the last layer of the model adapted on the Caltech101 dataset. For PVeRA, we take the learned $\boldsymbol \mu_q$ and $\boldsymbol \sigma_q$, and for VeRA, we take the equivalent to $\boldsymbol \mu_q$, i.e., the output of $\boldsymbol x \boldsymbol A_q \odot \boldsymbol d_q$. First, different draws from the $\mathcal N (\boldsymbol \mu_q, \boldsymbol \sigma^2_q)$ distributions have an impact of the latent adaptation, indicating that learning a distribution from which we sample for training acts as a sort of latent augmentation. We can conclude that the latent augmentations are beneficial for training seeing the overall better results from PVeRA over VeRA in Table~\ref{tab:main}. Moreover, we notice that the different latent adaptations of PVeRA seem to put more focus on semantically relevant parts of the image (e.g., the head of the elephant, the key features of the face) than for VeRA.

\subsection{Ablations}
We explore the impact of the rank and the position of the adapter on the final result. In order to limit computationally expensive experiments, we run the ablations across six randomly chosen datasets, two from each dataset category (DTD, CIFAR-100, Camelyon, Retinopathy, Clevr-Dist, and KITTI-dist), and report the average best validation loss across three seeds.

\mypar{Rank ablation.} In Table~\ref{tab:rank_ablation}, we test four values of the rank and look at the impact on the average best validation loss and the number of trainable parameters. We find that a rank of $r=256$, even though only slighly better than $r\in\{128,~512\}$, gives the best validation loss.
\begin{table}[]
    \centering
    \begin{tabular}{l c c c c}
        \toprule
        Rank & 64 & 128 & 256 & 512 \\
        Loss & 1.875 & 1.821 & 1.819 & 1.824 \\
        \# Parameters & 21 504 & 24 576 & 30 720 & 43 008\\
        \bottomrule
    \end{tabular}
    \caption{\textbf{Impact of the rank of PVeRA}. Validation loss and number of trainable parameters as a function of the adapter rank. Experiments ran for PVeRA on $Q$ and $V$.}
    \label{tab:rank_ablation}
\end{table}

\mypar{Architecture ablation.} We tested six architecture for the PVeRA adapter, putting adapters on variations of the query, key, and value components of the attention mechanism. In Table~\ref{tab:position_ablation}, we show that the best validation loss is obtained when applying PVeRA on the query and value components, similarly to the original LoRA paper. Because of how the attention is computed, we did not consider applying the adapter of the key and value branch.
\begin{table}[]
    \centering
    \begin{tabular}{c c c c c}
        \toprule
        $Q$ & $K$ & $V$ & Loss & \# Parameters\\
        \midrule
        \checkmark & & & 1.825 & 15 360 \\
        & \checkmark & & 1.827 & 15 360 \\
        & & \checkmark & 1.826 & 15 360 \\
        \checkmark & \checkmark & & 1.835 & 30 720 \\
        \checkmark & & \checkmark & 1.819 & 30 720 \\
        \checkmark & \checkmark & \checkmark & 1.821 & 46 080 \\
        \bottomrule
    \end{tabular}
    \caption{\textbf{Impact of the position of PVeRA}. Validation loss and number of trainable parameters of the PVeRA adapter for different positions on the query, key, and value elements of the attention mechanism. Experiments ran with $r=256$.}
    \label{tab:position_ablation}
\end{table}
\section{Conclusion}
We have introduced PVeRA, a probabilistic variant of VeRA. We demonstrated that our proposed PVeRA adapter outperforms the VeRA adapter on which it is based, all while showing the following qualities: conserved calibration, ability to generate confidence intervals for predictions, possibility to detect out-of-distribution samples. While we have proven the superiority of PVeRA over VeRA in the context of image classification, and have begun exploring its performance in natural language processing, we have not yet explored its performance on other computer vision tasks (e.g., adapting SAM~\cite{kirillov2023segany} for segmentating specific image types such as medical images). Future extensions of PVeRA could delve into these unexplored aspects, as well as modeling different distributions or other uncertainty formulations.

\mypar{Acknowledgments.} This work has benefited from state financial aid, managed by the Agence Nationale de Recherche under the investment program integrated into France 2030, project references ANR-21-RHUS-0003, ANR-23-IAHU-0002, and ANR-23-IACL-0003 – DATAIA CLUSTER (as part of IA CLUSTER program). This work was granted access to the HPC resources of IDRIS under the allocation 2024-AD011014802R1 made by GENCI.

{
    \small
    \bibliographystyle{ieeenat_fullname}
    \bibliography{main}

@String(NIPS= {Adv. Neural Inform. Process. Syst.})

@String(ICLR = {Int. Conf. Learn. Represent.})

@String(AAAI = {AAAI})

@String(NIPS  = {NeurIPS})

@String(ICLR  = {ICLR})

@misc{zhai2020vtab1k,
      title={A Large-scale Study of Representation Learning with the Visual Task Adaptation Benchmark}, 
      author={Xiaohua Zhai and Joan Puigcerver and Alexander Kolesnikov and Pierre Ruyssen and Carlos Riquelme and Mario Lucic and Josip Djolonga and Andre Susano Pinto and Maxim Neumann and Alexey Dosovitskiy and Lucas Beyer and Olivier Bachem and Michael Tschannen and Marcin Michalski and Olivier Bousquet and Sylvain Gelly and Neil Houlsby},
      year={2020},
      eprint={1910.04867},
      archivePrefix={arXiv},
      primaryClass={cs.CV},
      url={https://arxiv.org/abs/1910.04867}, 
}

@inproceedings{houlsby2019parameter,
  title = {Parameter-Efficient Transfer Learning for {NLP}},
  author = {Houlsby, Neil and Giurgiu, Andrei and Jastrzebski, Stanislaw and Morrone, Bruna and De Laroussilhe, Quentin and Gesmundo, Andrea and Attariyan, Mona and Gelly, Sylvain},
  booktitle = {Proceedings of the 36th International Conference on Machine Learning},
  year = {2019},
}

@misc{chen2022adaptformer,
      title={AdaptFormer: Adapting Vision Transformers for Scalable Visual Recognition}, 
      author={Shoufa Chen and Chongjian Ge and Zhan Tong and Jiangliu Wang and Yibing Song and Jue Wang and Ping Luo},
      year={2022},
      eprint={2205.13535},
      archivePrefix={arXiv},
      primaryClass={cs.CV},
      url={https://arxiv.org/abs/2205.13535}, 
}

@misc{bahdanau2016neural,
      title={Neural Machine Translation by Jointly Learning to Align and Translate}, 
      author={Dzmitry Bahdanau and Kyunghyun Cho and Yoshua Bengio},
      year={2016},
      eprint={1409.0473},
      archivePrefix={arXiv},
      primaryClass={cs.CL}
}

@inproceedings{ashish2017attention,
 author = {Vaswani, Ashish and Shazeer, Noam and Parmar, Niki and Uszkoreit, Jakob and Jones, Llion and Gomez, Aidan N and Kaiser, \L ukasz and Polosukhin, Illia},
 booktitle = {Advances in Neural Information Processing Systems},
 editor = {I. Guyon and U. Von Luxburg and S. Bengio and H. Wallach and R. Fergus and S. Vishwanathan and R. Garnett},
 pages = {},
 publisher = {Curran Associates, Inc.},
 title = {Attention is All you Need},
 url = {https://proceedings.neurips.cc/paper_files/paper/2017/file/3f5ee243547dee91fbd053c1c4a845aa-Paper.pdf},
 volume = {30},
 year = {2017}
}

@inproceedings{
    hu2022lora,
    title={Lo{RA}: Low-Rank Adaptation of Large Language Models},
    author={Edward J Hu and Yelong Shen and Phillip Wallis and Zeyuan Allen-Zhu and Yuanzhi Li and Shean Wang and Lu Wang and Weizhu Chen},
    booktitle={International Conference on Learning Representations},
    year={2022},
    url={https://openreview.net/forum?id=nZeVKeeFYf9}
}

@inproceedings{
    He2025SMT,
    title={SMT: Fine-Tuning Large Language Models with Sparse Matrices},
    author={He, Haoze and Li, Juncheng and Jiang, Xuan and Miller, Heather},
    booktitle={International Conference on Learning Representations},
    year={2025},
    url={https://openreview.net/forum?id=GbgCRJedQ7}
}

@inproceedings{kopiczko2024vera,
title={Ve{RA}: Vector-based Random Matrix Adaptation},
author={Dawid Jan Kopiczko and Tijmen Blankevoort and Yuki M Asano},
booktitle={The Twelfth International Conference on Learning Representations},
year={2024},
url={https://openreview.net/forum?id=NjNfLdxr3A}
}

@inproceedings{platt1999probabilistic,
  author = {Platt, J.},
  booktitle = {Advances in Large Margin Classifiers},
  title = {Probabilistic outputs for support vector machines and comparison to regularized
    likelihood methods},
  year = 1999
}

@inproceedings{zadrozny2001obtaining,
  title = {Obtaining calibrated probability estimates from decision trees and naive Bayesian classifiers},
  author = {Bianca Zadrozny and Charles Elkan},
  year = {2001},
  researchr = {https://researchr.org/publication/ZadroznyE01},
  cites = {0},
  citedby = {0},
  pages = {609-616},
  booktitle = {Proceedings of the Eighteenth International Conference on Machine Learning (ICML 2001), Williams College, Williamstown, MA, USA, June 28 - July 1, 2001},
  editor = {Carla E. Brodley and Andrea Pohoreckyj Danyluk},
  publisher = {Morgan Kaufmann},
  isbn = {1-55860-778-1},
}

@misc{guo2017calibration,
      title={On Calibration of Modern Neural Networks}, 
      author={Chuan Guo and Geoff Pleiss and Yu Sun and Kilian Q. Weinberger},
      year={2017},
      eprint={1706.04599},
      archivePrefix={arXiv},
      primaryClass={cs.LG}
}

@conference {naeini2015obtaining,
	title = {Obtaining Well Calibrated Probabilities Using Bayesian Binning.},
	booktitle = {AAAI},
	year = {2015},
	pages = {2901{\textendash}2907},
	author = {Mahdi Pakdaman Naeini and Cooper, Gregory F and Hauskrecht, Milos}
}

@misc{mukhoti2020calibrating,
      title={Calibrating Deep Neural Networks using Focal Loss}, 
      author={Jishnu Mukhoti and Viveka Kulharia and Amartya Sanyal and Stuart Golodetz and Philip H. S. Torr and Puneet K. Dokania},
      year={2020},
      eprint={2002.09437},
      archivePrefix={arXiv},
      primaryClass={cs.LG}
}

@misc{oquab2023dinov2,
  title={DINOv2: Learning Robust Visual Features without Supervision},
  author={Oquab, Maxime and Darcet, Timothée and Moutakanni, Theo and Vo, Huy V. and Szafraniec, Marc and Khalidov, Vasil and Fernandez, Pierre and Haziza, Daniel and Massa, Francisco and El-Nouby, Alaaeldin and Howes, Russell and Huang, Po-Yao and Xu, Hu and Sharma, Vasu and Li, Shang-Wen and Galuba, Wojciech and Rabbat, Mike and Assran, Mido and Ballas, Nicolas and Synnaeve, Gabriel and Misra, Ishan and Jegou, Herve and Mairal, Julien and Labatut, Patrick and Joulin, Armand and Bojanowski, Piotr},
  journal={arXiv:2304.07193},
  year={2023}
}

@misc{chen2020simple,
      title={A Simple Framework for Contrastive Learning of Visual Representations}, 
      author={Ting Chen and Simon Kornblith and Mohammad Norouzi and Geoffrey Hinton},
      year={2020},
      eprint={2002.05709},
      archivePrefix={arXiv},
      primaryClass={cs.LG}
}

@misc{liu2022ia3,
      title={Few-Shot Parameter-Efficient Fine-Tuning is Better and Cheaper than In-Context Learning}, 
      author={Haokun Liu and Derek Tam and Mohammed Muqeeth and Jay Mohta and Tenghao Huang and Mohit Bansal and Colin Raffel},
      year={2022},
      eprint={2205.05638},
      archivePrefix={arXiv},
      primaryClass={cs.LG},
      url={https://arxiv.org/abs/2205.05638}, 
}

@misc{radford2021learning,
      title={Learning Transferable Visual Models From Natural Language Supervision}, 
      author={Alec Radford and Jong Wook Kim and Chris Hallacy and Aditya Ramesh and Gabriel Goh and Sandhini Agarwal and Girish Sastry and Amanda Askell and Pamela Mishkin and Jack Clark and Gretchen Krueger and Ilya Sutskever},
      year={2021},
      eprint={2103.00020},
      archivePrefix={arXiv},
      primaryClass={cs.CV}
}

@article{kirillov2023segany,
  title={Segment Anything},
  author={Kirillov, Alexander and Mintun, Eric and Ravi, Nikhila and Mao, Hanzi and Rolland, Chloe and Gustafson, Laura and Xiao, Tete and Whitehead, Spencer and Berg, Alexander C. and Lo, Wan-Yen and Doll{\'a}r, Piotr and Girshick, Ross},
  journal={arXiv:2304.02643},
  year={2023}
}

@misc{yosinski2014transferable,
      title={How transferable are features in deep neural networks?}, 
      author={Jason Yosinski and Jeff Clune and Yoshua Bengio and Hod Lipson},
      year={2014},
      eprint={1411.1792},
      archivePrefix={arXiv},
      primaryClass={cs.LG}
}

@misc{ba2016layer,
  author = {Ba, Jimmy Lei and Kiros, Jamie Ryan and Hinton, Geoffrey E.},
  description = {[1607.06450] Layer Normalization},
  note = {cite arxiv:1607.06450},
  title = {Layer Normalization},
  url = {http://arxiv.org/abs/1607.06450},
  year = 2016
}

@misc{mahabadi2021compacter,
      title={Compacter: Efficient Low-Rank Hypercomplex Adapter Layers}, 
      author={Rabeeh Karimi Mahabadi and James Henderson and Sebastian Ruder},
      year={2021},
      eprint={2106.04647},
      archivePrefix={arXiv},
      primaryClass={cs.CL}
}

@misc{elizalde2022clap,
      title={CLAP: Learning Audio Concepts From Natural Language Supervision}, 
      author={Benjamin Elizalde and Soham Deshmukh and Mahmoud Al Ismail and Huaming Wang},
      year={2022},
      eprint={2206.04769},
      archivePrefix={arXiv},
      primaryClass={cs.SD},
      url={https://arxiv.org/abs/2206.04769}, 
}

@article{zhou2022learning,
   title={Learning to Prompt for Vision-Language Models},
   volume={130},
   ISSN={1573-1405},
   url={http://dx.doi.org/10.1007/s11263-022-01653-1},
   DOI={10.1007/s11263-022-01653-1},
   number={9},
   journal={International Journal of Computer Vision},
   publisher={Springer Science and Business Media LLC},
   author={Zhou, Kaiyang and Yang, Jingkang and Loy, Chen Change and Liu, Ziwei},
   year={2022},
   month=jul, pages={2337–2348} }

@misc{yao2023visual,
      title={Visual-Language Prompt Tuning with Knowledge-guided Context Optimization}, 
      author={Hantao Yao and Rui Zhang and Changsheng Xu},
      year={2023},
      eprint={2303.13283},
      archivePrefix={arXiv},
      primaryClass={cs.CV},
      url={https://arxiv.org/abs/2303.13283}, 
}

@misc{lester2021power,
      title={The Power of Scale for Parameter-Efficient Prompt Tuning}, 
      author={Brian Lester and Rami Al-Rfou and Noah Constant},
      year={2021},
      eprint={2104.08691},
      archivePrefix={arXiv},
      primaryClass={cs.CL}
}

@misc{jia2022visual,
      title={Visual Prompt Tuning}, 
      author={Menglin Jia and Luming Tang and Bor-Chun Chen and Claire Cardie and Serge Belongie and Bharath Hariharan and Ser-Nam Lim},
      year={2022},
      eprint={2203.12119},
      archivePrefix={arXiv},
      primaryClass={cs.CV}
}

@inproceedings{sparse2024bhardwaj,
 author = {Bhardwaj, Kartikeya and Pandey, Nilesh Prasad and Priyadarshi, Sweta and Ganapathy, Viswanath and Kadambi, Shreya and Esteves, Rafael and Borse, Shubhankar and Whatmough, Paul and Garrepalli, Risheek and Van Baalen, Mart and Teague, Harris and Nagel, Markus},
 booktitle = {Advances in Neural Information Processing Systems},
 editor = {A. Globerson and L. Mackey and D. Belgrave and A. Fan and U. Paquet and J. Tomczak and C. Zhang},
 pages = {13685--13715},
 publisher = {Curran Associates, Inc.},
 title = {Sparse High Rank Adapters},
 url = {https://proceedings.neurips.cc/paper_files/paper/2024/file/18c0102cb7f1a02c14f0929089b2e576-Paper-Conference.pdf},
 volume = {37},
 year = {2024}
}

@article{wang2024universality,
  title={Universality and limitations of prompt tuning},
  author={Wang, Yihan and Chauhan, Jatin and Wang, Wei and Hsieh, Cho-Jui},
  journal={Advances in Neural Information Processing Systems},
  volume={36},
  year={2024}
}

@inproceedings{kingma2014vae,
  author = {Kingma, Diederik P. and Welling, Max},
  booktitle = {2nd International Conference on Learning Representations, {ICLR} 2014, Banff, AB, Canada, April 14-16, 2014, Conference Track Proceedings},
  eprint = {http://arxiv.org/abs/1312.6114v10},
  eprintclass = {stat.ML},
  eprinttype = {arXiv},
  title = {{Auto-Encoding Variational Bayes}},
  year = 2014
}

@book{radford1996bayesian,
author = {Neal, Radford M.},
title = {Bayesian Learning for Neural Networks},
year = {1996},
isbn = {0387947248},
publisher = {Springer-Verlag},
address = {Berlin, Heidelberg}
}

@article{mackay1992bnn,
    author = {MacKay, David J. C.},
    title = "{A Practical Bayesian Framework for Backpropagation Networks}",
    journal = {Neural Computation},
    volume = {4},
    number = {3},
    pages = {448-472},
    year = {1992},
    month = {05},
    issn = {0899-7667},
    doi = {10.1162/neco.1992.4.3.448},
    url = {https://doi.org/10.1162/neco.1992.4.3.448},
    eprint = {https://direct.mit.edu/neco/article-pdf/4/3/448/812348/neco.1992.4.3.448.pdf},
}

@misc{gal2016dropout,
      title={Dropout as a Bayesian Approximation: Representing Model Uncertainty in Deep Learning}, 
      author={Yarin Gal and Zoubin Ghahramani},
      year={2016},
      eprint={1506.02142},
      archivePrefix={arXiv},
      primaryClass={stat.ML}
}

@article{oh2018modeling,
  title={Modeling uncertainty with hedged instance embedding},
  author={Oh, Seong Joon and Murphy, Kevin and Pan, Jiyan and Roth, Joseph and Schroff, Florian and Gallagher, Andrew},
  journal={arXiv preprint arXiv:1810.00319},
  year={2018}
}

@inproceedings{chun2021probabilistic,
  title={Probabilistic embeddings for cross-modal retrieval},
  author={Chun, Sanghyuk and Oh, Seong Joon and De Rezende, Rafael Sampaio and Kalantidis, Yannis and Larlus, Diane},
  booktitle={Proceedings of the IEEE/CVF Conference on Computer Vision and Pattern Recognition},
  pages={8415--8424},
  year={2021}
}

@inproceedings{shi2019probabilistic,
  title={Probabilistic face embeddings},
  author={Shi, Yichun and Jain, Anil K},
  booktitle={Proceedings of the IEEE/CVF International Conference on Computer Vision},
  pages={6902--6911},
  year={2019}
}

@misc{upadhyay2023probvlm,
      title={ProbVLM: Probabilistic Adapter for Frozen Vision-Language Models}, 
      author={Uddeshya Upadhyay and Shyamgopal Karthik and Massimiliano Mancini and Zeynep Akata},
      year={2023},
      eprint={2307.00398},
      archivePrefix={arXiv},
      primaryClass={cs.CV}
}

@article{liu2024dora,
  title={DoRA: Weight-Decomposed Low-Rank Adaptation},
  author={Liu, Shih-Yang and Wang, Chien-Yi and Yin, Hongxu and Molchanov, Pavlo and Wang, Yu-Chiang Frank and Cheng, Kwang-Ting and Chen, Min-Hung},
  journal={arXiv preprint arXiv:2402.09353},
  year={2024}
}

@article{li2006one,
  title={One-shot learning of object categories},
  author={Li, Fei-Fei and Fergus, Rob and Perona, Pietro},
  journal={IEEE Transactions on Pattern Analysis and Machine Intelligence},
  year={2006},
  publisher={IEEE}
}

@TECHREPORT{cifar10,
    author = {Alex Krizhevsky},
    title = {Learning multiple layers of features from tiny images},
    institution = {},
    year = {2009}
}

@InProceedings{cimpoi14describing,
	      Author    = {M. Cimpoi and S. Maji and I. Kokkinos and S. Mohamed and and A. Vedaldi},
	      Title     = {Describing Textures in the Wild},
	      Booktitle = {IEEE Conference on Computer Vision and Pattern Recognition},
	      Year      = {2014}
}

@InProceedings{Nilsback08, 
   author = "Nilsback, M-E. and Zisserman, A.", 
   title = "Automated Flower Classification over a Large Number of Classes", 
   booktitle = "Indian Conference on Computer Vision, Graphics and Image Processing", 
   year = "2008", 
   month = "Dec" 
}

@InProceedings{parkhi12a,
  author       = "Parkhi, O. M. and Vedaldi, A. and Zisserman, A. and Jawahar, C.~V.",
  title        = "Cats and Dogs",
  booktitle    = "IEEE Conference on Computer Vision and Pattern Recognition",
  year         = "2012",
}

@inproceedings{xiao2010sun,
  title={Sun database: Large-scale scene recognition from abbey to zoo},
  author={Xiao, Jianxiong and Hays, James and Ehinger, Krista A and Oliva, Aude and Torralba, Antonio},
  booktitle={IEEE Conference on Computer Vision and Pattern Recognition},
  year={2010}
}

@inproceedings{netzer2011reading,
title	= {Reading Digits in Natural Images with Unsupervised Feature Learning},
author	= {Yuval Netzer and Tao Wang and Adam Coates and Alessandro Bissacco and Bo Wu and Andrew Y. Ng},
year	= {2011},
booktitle	= {NIPS Workshop on Deep Learning and Unsupervised Feature Learning 2011}
}

@inproceedings{veeling2018rotation,
  title={Rotation equivariant cnns for digital pathology},
  author={Veeling, Bastiaan S and Linmans, Jasper and Winkens, Jim and Cohen, Taco and Welling, Max},
  booktitle={International Conference on Medical Image Computing and Computer-Assisted Intervention},
  year={2018},
}

@article{helber2017eurosat,
  title={Eurosat: A novel dataset and deep learning benchmark for land use and land cover classification},
  author={Helber, Patrick and Bischke, Benjamin and Dengel, Andreas and Borth, Damian},
  journal={IEEE Journal of Selected Topics in Applied Earth Observations and Remote Sensing},
  year={2019},
}

@article{cheng2017remote,
  title={Remote sensing image scene classification: Benchmark and state of the art},
  author={Cheng, Gong and Han, Junwei and Lu, Xiaoqiang},
  journal={Proceedings of the IEEE},
  year={2017},
  publisher={IEEE}
}

@misc{kaggle-diabetic-retinopathy,
    author = "Kaggle and EyePacs",
    title  = "Kaggle Diabetic Retinopathy Detection",
    month  = "July",
    year   = "2015",
    url    = "https://www.kaggle.com/c/diabetic-retinopathy-detection/data"
}

@inproceedings{johnson2017clevr,
  title={Clevr: A diagnostic dataset for compositional language and elementary visual reasoning},
  author={Johnson, Justin and Hariharan, Bharath and van der Maaten, Laurens and Fei-Fei, Li and Lawrence Zitnick, C and Girshick, Ross},
  booktitle={IEEE Conference on Computer Vision and Pattern Recognition},
  year={2017}
}

@article{beattie2016deepmind,
  title={Deepmind lab},
  author={Beattie, Charles and Leibo, Joel Z and Teplyashin, Denis and Ward, Tom and Wainwright, Marcus and K{\"u}ttler, Heinrich and Lefrancq, Andrew and Green, Simon and Vald{\'e}s, V{\'\i}ctor and Sadik, Amir and others},
  journal={arXiv preprint arXiv:1612.03801},
  year={2016}
}

@misc{dsprites17,
author = {Loic Matthey and Irina Higgins and Demis Hassabis and Alexander Lerchner},
title = {dSprites: Disentanglement testing Sprites dataset},
howpublished= {https://github.com/deepmind/dsprites-dataset/},
year = "2017",
}

@ARTICLE{Geiger2013IJRR,
  author = {Andreas Geiger and Philip Lenz and Christoph Stiller and Raquel Urtasun},
  title = {Vision meets Robotics: The KITTI Dataset},
  journal = {International Journal of Robotics Research},
  year = {2013}
}

@inproceedings{lecun2004learning,
  title={Learning methods for generic object recognition with invariance to pose and lighting},
  author={LeCun, Yann and Huang, Fu Jie and Bottou, Leon},
  booktitle={IEEE Conference on Computer Vision and Pattern Recognition},
  year={2004}
}

@misc{loshchilov2019decoupledweightdecayregularization,
      title={Decoupled Weight Decay Regularization}, 
      author={Ilya Loshchilov and Frank Hutter},
      year={2019},
      eprint={1711.05101},
      archivePrefix={arXiv},
      primaryClass={cs.LG},
      url={https://arxiv.org/abs/1711.05101}, 
}

@misc{he2023debertav3improvingdebertausing,
      title={DeBERTaV3: Improving DeBERTa using ELECTRA-Style Pre-Training with Gradient-Disentangled Embedding Sharing}, 
      author={Pengcheng He and Jianfeng Gao and Weizhu Chen},
      year={2023},
      eprint={2111.09543},
      archivePrefix={arXiv},
      primaryClass={cs.CL},
      url={https://arxiv.org/abs/2111.09543}, 
}

@misc{wang2019gluemultitaskbenchmarkanalysis,
      title={GLUE: A Multi-Task Benchmark and Analysis Platform for Natural Language Understanding}, 
      author={Alex Wang and Amanpreet Singh and Julian Michael and Felix Hill and Omer Levy and Samuel R. Bowman},
      year={2019},
      eprint={1804.07461},
      archivePrefix={arXiv},
      primaryClass={cs.CL},
      url={https://arxiv.org/abs/1804.07461}, 
}
}

\clearpage
\setcounter{page}{1}
\renewcommand{\thesection}{\Alph{section}}
\setcounter{section}{0}
\renewcommand*{\theHsection}{chX.\the\value{section}}
\maketitlesupplementary

\section{Grid Search}
\label{supp_sec:grid_search}
Table~\ref{tab:grid_search} shows the values of grid search used for the rank, the reduction ratio, and the learning rate, as well as the percentage of configurations for which each value performed best.
\begin{table}[b]
    \newcolumntype{C}{>{\centering\arraybackslash}p{1.5cm}}
    \newcolumntype{L}{>{\arraybackslash}p{1.3cm}}
    \centering
    
    \begin{subtable}[b]{\linewidth}
    \centering
    \begin{tabular}{L C C C}
    \toprule
        & 32 & 64 & 128 \\
    \midrule
        LoRA & 26\% & 35\% & 39\% \\
        DoRA & 18\% & 37\% & 46\% \\
    \bottomrule
    \end{tabular}
    \caption{Rank}
    \end{subtable}
    
    \vspace{0.3cm}
    \begin{subtable}[b]{\linewidth}
    \centering
    \begin{tabular}{L C C C}
    \toprule
        & 16 & 8 & 4 \\
    \midrule
        Bottleneck & 11\% & 21\% & 68\% \\
        AdaptFormer & 11\% & 23\% & 67\% \\
    \bottomrule
    \end{tabular}
    \caption{Reduction ratio}
    \end{subtable}

    \vspace{0.3cm}
    \begin{subtable}[b]{\linewidth}
    \centering
    \begin{tabular}{L C C C}
    \toprule
        & $10^{-3}$ & $10^{-4}$ & $10^{-5}$ \\
    \midrule
        Linear & 79\% & 21\% & 0\%\\
        (IA)$^3$ & 60\% & 30\% & 10\% \\
        VeRA & 46\% & 23\% & 32\% \\
        PVeRA &  67\% & 28\% & 5\% \\
    \bottomrule
    \end{tabular}
    \caption{Learning rate}
    \end{subtable}
    
    \caption{\textbf{Grid search results}. Grid search hyperparameters for (a) ranks, (b) reduction ratios, and (c) learning rates, along with the percentage of the 57 configurations (19 datasets across three seeds) for which each value was chosen.}
    \label{tab:grid_search}
\end{table}
\section{Pseudocode for PVeRA}
\label{supp_sec:pseudocode}
Algorithm~\ref{alg:pvera} shows pseudocode for the initialization and the forward pass through PVeRA. Our Python code is available \href{https://github.com/leofillioux/pvera}{here}.

\begin{algorithm}
\caption{\textbf{Initialization and forward pass of PVeRA}.
Algorithm for the initialization and forward pass of PVeRA.}\label{alg:pvera}
\begin{algorithmic}
\Input $\boldsymbol x$, $\alpha$, $\boldsymbol A$, $\boldsymbol B$, linear\_layer, training, $r$, $d$

\Init
\State $\boldsymbol b \gets$ zeros($d$)
\State $\boldsymbol d \gets$ ones($2r$)$\cdot$uniform($10^{-5}, 1$)

\Forward

\State $\boldsymbol b \gets$ repeat($\boldsymbol b$, $\boldsymbol x$.batch\_size, $\boldsymbol x$.seq\_length)
\State $\boldsymbol d \gets$ repeat($\boldsymbol d$, $\boldsymbol x$.batch\_size, $\boldsymbol x$.seq\_length)

\State $\boldsymbol \mu$, $\boldsymbol \sigma \gets \boldsymbol x \cdot \boldsymbol A \odot \boldsymbol d$

\If{training}

\State $\boldsymbol \epsilon \gets $ random\_normal($\boldsymbol \mu$.shape)
\State $\boldsymbol z \gets \alpha \cdot (\boldsymbol \mu + \boldsymbol \epsilon \odot \exp (\boldsymbol \sigma^2))$

\Else

\State $\boldsymbol z \gets \boldsymbol \mu$
\EndIf

\State $\boldsymbol x_\text{adapted} \gets \alpha \cdot \boldsymbol z \odot \boldsymbol b$

\Returns
\State linear\_layer($\boldsymbol x$) + $\boldsymbol x_\text{adapted}$

\end{algorithmic}
\end{algorithm}
\section{Uncertainty Estimation}
\label{supp_sec:uncertainty}

\begin{figure}
    \centering
    \includegraphics[width=0.5\linewidth]{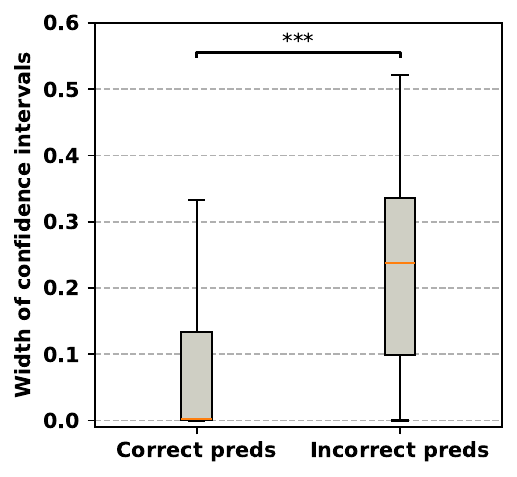}
    \caption{\textbf{Width of confidence intervals for correctly and incorrectly classified samples}. Width ($(\text{upper bound}) - (\text{lower bound})$) for correctly classified and incorrectly classified samples across the test set from all VTAB-1k datasets, generated using Monte Carlo inference with random sampling of the PVeRA adapters. The significance level corresponds to the p-value of a one-sided unpaired Wilcoxon test.}
    \label{fig:mc_confint}
\end{figure}

\begin{figure*}
    \centering
    \begin{subfigure}{0.97\linewidth}
        \centering
        \includegraphics[width=\textwidth]{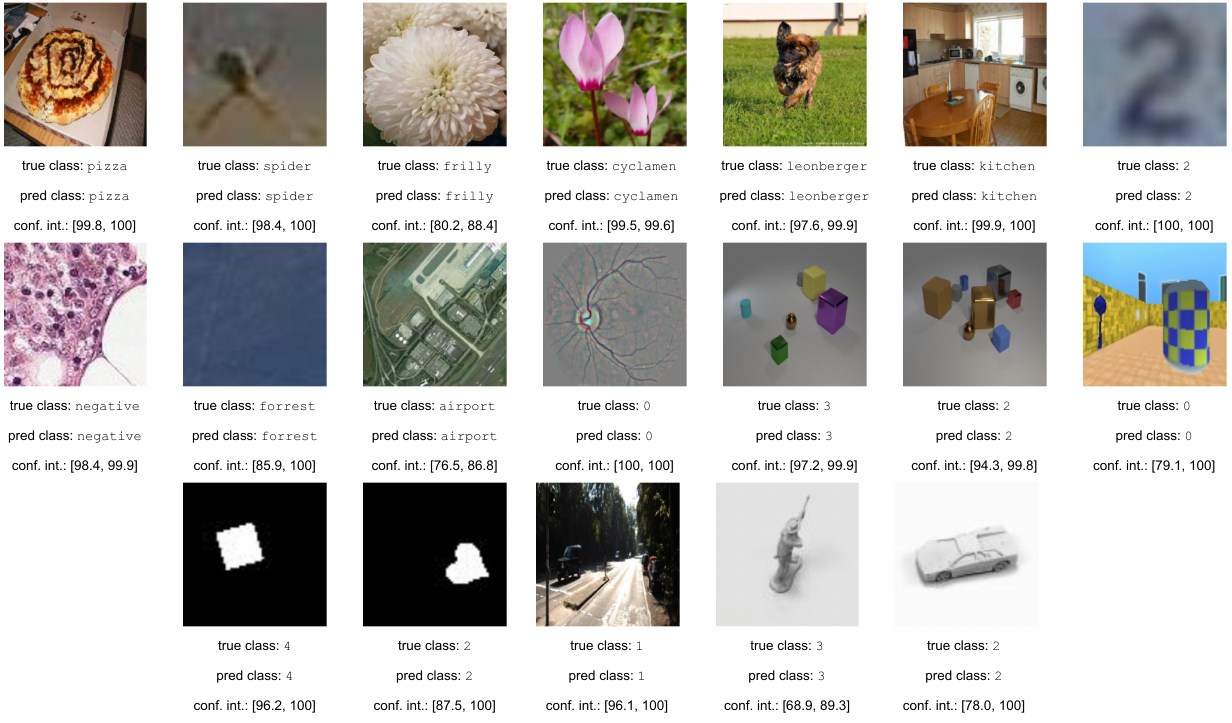}
        \caption{Correctly classified}
    \end{subfigure}
    
    \vspace{0.3cm}
    \begin{subfigure}{\linewidth}
        \centering
        \includegraphics[width=0.97\textwidth]{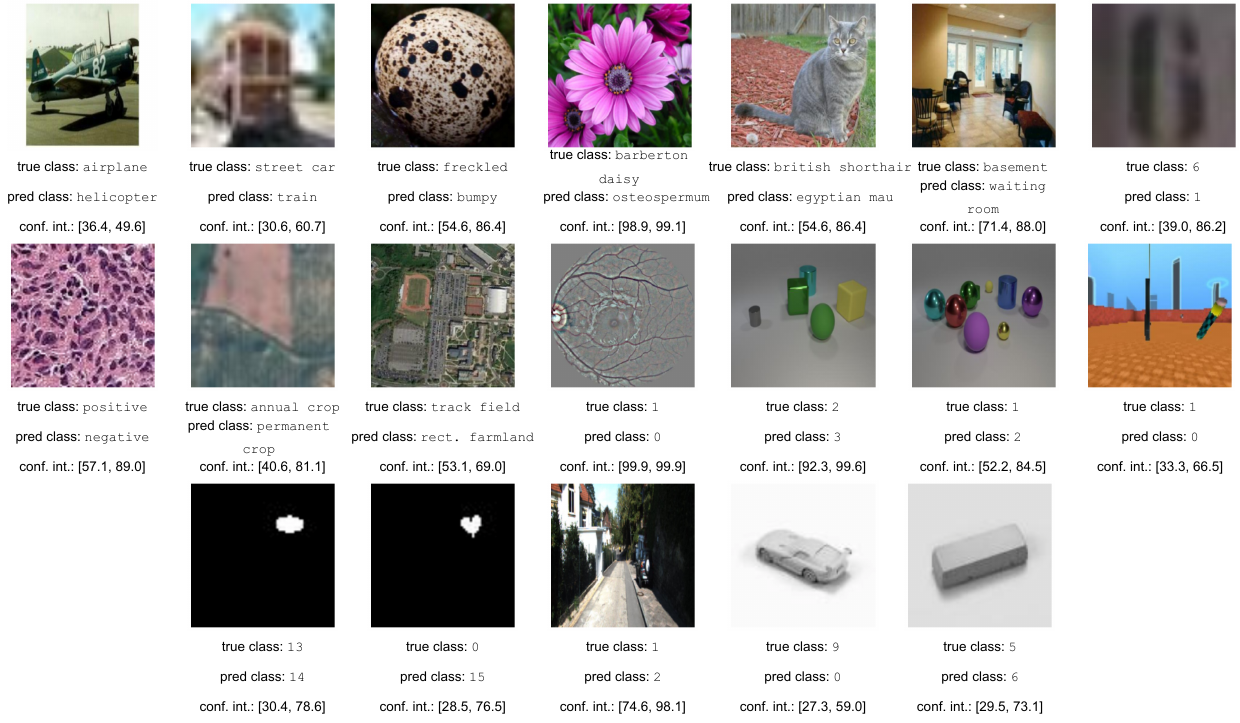}
        \caption{Incorrectly classified}
    \end{subfigure}
    \caption{\textbf{Confidence interval estimation}. True class, predicted class, and estimated 95\% confidence intervals for all 19 datasets (a) correctly classified samples, and (b) incorrectly classified samples.}
    \label{fig:supp_uncertainty}
\end{figure*}

Sampling from the learned distribution during inference allows to generate confidence intervals for the predictions by making multiple consecutive passes with each sample. In Figure~\ref{fig:supp_uncertainty}, we look at each dataset of VTAB-1k and analyze images that were correctly classified and incorrectly classified. The predicted class is the most predicted class among the $k$ predictions, and the confidence intervals is computed from the softmax scores of this predicted class. Figure~\ref{fig:mc_confint} shows the widths of the Monte Carlo confidence intervals for the correctly and incorrectly classified samples across all datasets. It is noteworthy to observe that the confidence intervals of the wrongly classified images is much larger than those of the correctly classified images. Moreover, even when an image is correctly classified, the spread of the confidence interval can indicate uncertainty. For example in Figure~\ref{fig:supp_uncertainty}.a, EuroSAT's image of the \texttt{forest} could have easily been classified \texttt{river} or \texttt{sea \& lake}, which explains the higher confidence interval than Caltech101's image of \texttt{pizza} for which the uncertainty is much lower.

\section{Application on Natural Language}
\label{supp_sec:nlp}
While the main focus of the study is to introduce PVeRA on vision tasks, we include here a small experiment on adapting language models for different tasks. We utilize a DeBERTa-V3-base~\cite{he2023debertav3improvingdebertausing} model, which we adapt to three tasks from the GLUE benchmark~\cite{wang2019gluemultitaskbenchmarkanalysis} (we select the tasks with a low number of samples for reasons of computational efficiency). Table~\ref{tab:nlp_results} summarizes these results. Even across language classification tasks, PVeRA is very competitive compared to other adapters. While this is not an in-depth study, it shows very promising results. We train across three random seeds, we report results on the initial validation set, and split the provided train set into a training and validation set, as the test set labels are not made available. Note that AdaptFormer is not present as it was introduced for vision tasks and has, to the best of our knowledge, not been used for language tasks.
\begin{table}[h]
    \newcolumntype{C}{>{\centering\arraybackslash}p{1.5cm}}
    \newcolumntype{L}{>{\arraybackslash}p{1.7cm}}
    
    \centering
    \begin{tabular}{L C C C}
    \toprule
        Adapter & RTE & MRPC & CoLA \\
    \midrule
        Linear &
            $\;\,$63.2$^*$ &
            $\;\,$68.6$^*$ &
            $\;\,$48.3$^*$ \\
        Bottleneck &
            $\;\,$68.6$^*$ &
            85.5 &
            $\;\,$43.1$^*$ \\
        IA$^3$ &
            \underline{76.3} &
            85.6 &
            \underline{65.0} \\
        DoRA &
            $\;\,$71.8$^*$ &
            85.5 &
            62.9 \\
        LoRA &
            76.1 &
            85.4 &
            $\;\,$61.3$^*$ \\
        VeRA &
            $\;\,$72.8$^*$ &
            \textbf{85.9} &
            $\;\,$62.4$^*$ \\
        PVeRA &
        \textbf{76.7} &
        \underline{85.7} &
        \textbf{65.6} \\
    \bottomrule
    \end{tabular}
    
    \caption{\textbf{Natural language processing results}. Results on three classification tasks from the GLUE benchmark. All reported metrics are averaged over three random seeds. RTE and MRPC report accuracies, while CoLA reports the Matthews correlation coefficient. Results marked with $^*$ indicate accuracies significantly lower ($p=0.05$) than the accuracy of PVeRA.}
    \label{tab:nlp_results}
\end{table}
\section{Detailed Results}
\label{supp_sec:detailed_results}
Table~\ref{tab:detailed_results} presents a detailed version of Table~\ref{tab:main} with standard deviation of accuracies. We take the binary accuracy of each test set samples, perform a binomial test between each adapter and PVeRA, and mark the results with significantly lower accuracy with $^*$. Table~\ref{tab:detailed_results_small} presents similar results for a smaller DINOv2 (ViT-S/14). Conclusions are similar than for a ViT-B/14, with larger gaps in performance, with PVeRA's performance being $67.5\%$ with the second best performance being for LoRA and DoRA at $66.1\%$ and VeRA average performance at $65.9\%$. For a larger model, DINOv2 (ViT-L/14) presented in Table~\ref{tab:detailed_results_large}, the gap between the performance of PVeRA ($73.3\%$) and the second best (LoRA with $73.1\%$) is smaller. We believe that this is due to the fact that with larger models, the performance gets closer to the highest attainable performance, and the gains in performance are more difficult to obtain. Note that for computational efficiency constraints, we did not perform grid search for the ViT-L/14, and instead fixed the hyperparameters to the most chosen hyperparameter reported in Table~\ref{tab:grid_search} (e.g., rank of $128$ for LoRA, and learning rate of $10^{-3}$ for VeRA).
\begin{table*}[]
    \centering
    \resizebox{\textwidth}{!}{
    \begin{tabular}{l c c c c c c c c}
    \toprule
         & Linear & Bottleneck & IA$^3$ & AdaptFormer & DoRA & LoRA & VeRA & PVeRA \\
    \midrule
        \coloreddot{natural}~Caltech101 &
            85.1$^*$ $\pm$ 0.4$\;\,$ &
            88.6 $\pm$ 0.6 &
            86.8$^*$ $\pm$ 0.6$\;\,$ &
            87.9$^*$ $\pm$ 1.1$\;\,$ &
            90.0 $\pm$ 1.3 &
            90.1 $\pm$ 1.7 &
            87.3$^*$ $\pm$ 1.1$\;\,$ &
            88.7 $\pm$ 0.7 \\
        \coloreddot{natural}~CIFAR-100 & 
            61.6$^*$ $\pm$ 0.2$\;\,$ & 
            \:\:41.8$^*$ $\pm$ 11.4$\;\,$ &
            71.3 $\pm$ 0.8 & 
            56.8$^*$ $\pm$ 1.9$\;\,$ & 
            61.8$^*$ $\pm$ 0.8$\;\,$ & 
            64.5$^*$ $\pm$ 1.5$\;\,$ & 
            70.1$^*$ $\pm$ 0.0$\;\,$ & 
            71.7 $\pm$ 0.3 \\
        \coloreddot{natural}~DTD & 
            73.7$^*$ $\pm$ 0.3$\;\,$ & 
            72.3$^*$ $\pm$ 1.0$\;\,$ & 
            77.2 $\pm$ 0.5 & 
            72.9$^*$ $\pm$ 0.9$\;\,$ & 
            74.3$^*$ $\pm$ 0.2$\;\,$ & 
            75.3 $\pm$ 0.1 & 
            76.6 $\pm$ 0.3 & 
            76.1 $\pm$ 1.8 \\
        \coloreddot{natural}~Flowers102 & 
            99.7 $\pm$ 0.0 & 
            98.7$^*$ $\pm$ 0.4$\;\,$ & 
            99.7 $\pm$ 0.0 & 
            98.5$^*$ $\pm$ 0.2$\;\,$ & 
            99.1$^*$ $\pm$ 0.1$\;\,$ & 
            99.5$^*$ $\pm$ 0.0$\;\,$ & 
            99.7 $\pm$ 0.0 & 
            99.7 $\pm$ 0.0 \\
        \coloreddot{natural}~Pets & 
            94.0 $\pm$ 0.1 & 
            85.8$^*$ $\pm$ 7.5$\;\,$ & 
            94.1 $\pm$ 0.1 & 
            90.5$^*$ $\pm$ 0.2$\;\,$ & 
            90.7$^*$ $\pm$ 0.6$\;\,$ & 
            91.8$^*$ $\pm$ 0.4$\;\,$ & 
            94.1 $\pm$ 0.1 & 
            93.3 $\pm$ 0.9 \\
        \coloreddot{natural}~Sun397 & 
            51.3$^*$ $\pm$ 0.7$\;\,$ & 
            \:\:32.5$^*$ $\pm$ 21.7$\;\,$ & 
            54.2$^*$ $\pm$ 0.3$\;\,$ & 
            \:\:30.7$^*$ $\pm$ 20.4$\;\,$ & 
            49.1$^*$ $\pm$ 0.9$\;\,$ &
            50.7$^*$ $\pm$ 0.9$\;\,$ & 
            54.5$^*$ $\pm$ 0.2$\;\,$ & 
            54.7 $\pm$ 0.1 \\
        \coloreddot{natural}~SVHN & 
            38.9$^*$ $\pm$ 1.5$\;\,$ & 
            90.9 $\pm$ 0.8 & 
            68.7$^*$ $\pm$ 1.5$\;\,$ & 
            89.3 $\pm$ 0.8 & 
            91.0 $\pm$ 0.7 & 
            87.7$^*$ $\pm$ 0.4$\;\,$ & 
            87.8$^*$ $\pm$ 0.7$\;\,$ & 
            88.7 $\pm$ 0.3 \\

        \midrule

        \coloreddot{specialized}~Camelyon &
            82.3$^*$ $\pm$ 0.6$\;\,$ & 
            86.4 $\pm$ 0.7 & 
            83.2$^*$ $\pm$ 1.0$\;\,$ & 
            86.5 $\pm$ 1.4 & 
            87.2 $\pm$ 0.7 & 
            84.7$^*$ $\pm$ 0.1$\;\,$ & 
            84.3$^*$ $\pm$ 0.5$\;\,$ & 
            85.0 $\pm$ 0.9 \\
        \coloreddot{specialized}~EuroSAT &
            90.6$^*$ $\pm$ 0.2$\;\,$ & 
            94.3 $\pm$ 0.1 & 
            93.0$^*$ $\pm$ 0.3$\;\,$ & 
            94.3 $\pm$ 0.5 & 
            95.5 $\pm$ 0.3 & 
            94.9 $\pm$ 0.4 & 
            93.6$^*$ $\pm$ 0.4$\;\,$ & 
            94.3 $\pm$ 0.4 \\
        \coloreddot{specialized}~Resisc45 & 
            78.8$^*$ $\pm$ 0.2$\;\,$ & 
            82.7$^*$ $\pm$ 0.9$\;\,$ & 
            84.9$^*$ $\pm$ 0.5$\;\,$ & 
            83.6$^*$ $\pm$ 1.0$\;\,$ & 
            87.7 $\pm$ 0.5 & 
            84.9$^*$ $\pm$ 0.7$\;\,$ & 
            84.9$^*$ $\pm$ 0.2$\;\,$ & 
            86.4 $\pm$ 0.3 \\
        \coloreddot{specialized}~Retinopathy & 
            73.8$^*$ $\pm$ 0.3$\;\,$ & 
            73.6$^*$ $\pm$ 0.0$\;\,$ & 
            74.4$^*$ $\pm$ 1.1$\;\,$ & 
            73.6$^*$ $\pm$ 0.0$\;\,$ & 
            73.6$^*$ $\pm$ 0.0$\;\,$ & 
            75.0 $\pm$ 1.1 & 
            75.0 $\pm$ 0.1 & 
            74.8 $\pm$ 0.9 \\

        \midrule

        \coloreddot{structured}~Clevr-Count &
            42.9$^*$ $\pm$ 0.8$\;\,$ & 
            78.6 $\pm$ 1.0 & 
            55.9$^*$ $\pm$ 1.5$\;\,$ & 
            86.7 $\pm$ 0.7 & 
            63.7$^*$ $\pm$ 2.3$\;\,$ & 
            66.2$^*$ $\pm$ 1.8$\;\,$ & 
            58.1$^*$ $\pm$ 2.5$\;\,$ & 
            71.5 $\pm$ 4.8 \\
        \coloreddot{structured}~Clevr-Dist &
            33.7$^*$ $\pm$ 1.8$\;\,$ & 
            60.5 $\pm$ 0.7 & 
            51.8$^*$ $\pm$ 3.7$\;\,$ & 
            61.7 $\pm$ 1.0 & 
            60.5 $\pm$ 1.1 & 
            59.5$^*$ $\pm$ 1.1$\;\,$ & 
            58.5$^*$ $\pm$ 1.2$\;\,$ & 
            60.6 $\pm$ 0.4 \\
        \coloreddot{structured}~DMLab & 
            41.2$^*$ $\pm$ 0.8$\;\,$ & 
            48.1$^*$ $\pm$ 3.6$\;\,$ & 
            43.3$^*$ $\pm$ 0.5$\;\,$ & 
            49.3 $\pm$ 0.8 & 
            52.6 $\pm$ 1.6 & 
            50.3 $\pm$ 0.5 & 
            47.4$^*$ $\pm$ 1.2$\;\,$ & 
            48.6 $\pm$ 0.2 \\
        \coloreddot{structured}~dSpr-Loc & 
            11.8$^*$ $\pm$ 3.0$\;\,$ & 
            80.4 $\pm$ 2.7 & 
            55.5$^*$ $\pm$ 5.5$\;\,$ & 
            69.3$^*$ $\pm$ 4.0$\;\,$ & 
            84.3 $\pm$ 0.3 & 
            76.8 $\pm$ 0.5 & 
            73.8 $\pm$ 3.6 & 
            72.1 $\pm$ 1.5 \\
        \coloreddot{structured}~dSpr-Ori & 
            31.0$^*$ $\pm$ 2.8$\;\,$ & 
            49.9 $\pm$ 0.8 & 
            53.1 $\pm$ 0.8 & 
            52.2 $\pm$ 2.4 & 
            52.1 $\pm$ 0.2 & 
            52.3 $\pm$ 0.4 & 
            48.0$^*$ $\pm$ 0.7$\;\,$ & 
            49.5 $\pm$ 2.1 \\
        \coloreddot{structured}~KITTI-Dist & 
            54.0$^*$ $\pm$ 3.6$\;\,$ & 
            79.7$^*$ $\pm$ 2.3$\;\,$ & 
            77.7$^*$ $\pm$ 0.5$\;\,$ & 
            82.7 $\pm$ 2.0 & 
            82.1 $\pm$ 1.0 & 
            80.8$^*$ $\pm$ 2.4$\;\,$ & 
            84.0 $\pm$ 1.7 & 
            83.3 $\pm$ 0.4 \\
        \coloreddot{structured}~sNORB-Azim & 
            12.4$^*$ $\pm$ 0.5$\;\,$ & 
            19.1$^*$ $\pm$ 1.3$\;\,$ & 
            14.4$^*$ $\pm$ 1.7$\;\,$ & 
            19.6$^*$ $\pm$ 1.3$\;\,$ & 
            21.2 $\pm$ 1.0 & 
            19.6$^*$ $\pm$ 0.8$\;\,$ & 
            19.0$^*$ $\pm$ 0.4$\;\,$ & 
            20.3 $\pm$ 1.1 \\
        \coloreddot{structured}~sNORB-Elev & 
            25.5$^*$ $\pm$ 0.6$\;\,$ & 
            31.7$^*$ $\pm$ 2.5$\;\,$ & 
            26.8$^*$ $\pm$ 0.6$\;\,$ & 
            36.7 $\pm$ 1.0 & 
            32.2$^*$ $\pm$ 1.2$\;\,$ & 
            34.3$^*$ $\pm$ 2.1$\;\,$ & 
            32.3$^*$ $\pm$ 0.6$\;\,$ & 
            37.0 $\pm$ 2.8 \\
    \bottomrule
    \end{tabular}}
    \caption{\textbf{Detailed VTAB-1k benchmark results using a ViT-B/14}. Average and standard deviation accuracies across three seeds for seven adapters across all 19 VTAB-1k datasets. Results marked with $^*$ indicate accuracies significantly lower ($p=0.05$) than the accuracy of PVeRA.}
    \label{tab:detailed_results}
\end{table*}

\begin{table*}[]
    \centering
    \resizebox{\textwidth}{!}{
    \begin{tabular}{l c c c c c c c c}
    \toprule
         & Linear & Bottleneck & IA$^3$ & AdaptFormer & DoRA & LoRA & VeRA & PVeRA \\
    \midrule
        \coloreddot{natural}~Caltech101 &
            85.9$^*$ $\pm$ 0.4$\;\,$ &
            62.9$^*$ $\pm$ 2.1$\;\,$ &
            87.8 $\pm$ 1.9 &
            76.2$^*$ $\pm$ 1.8$\;\,$ &
            86.9$^*$ $\pm$ 2.1$\;\,$ &
            87.6$^*$ $\pm$ 1.1$\;\,$ &
            85.4$^*$ $\pm$ 0.4$\;\,$ &
            88.0 $\pm$ 0.2 \\
        \coloreddot{natural}~CIFAR-100 & 
            48.4$^*$ $\pm$ 0.3$\;\,$ &
            10.2$^*$ $\pm$ 3.3$\;\,$ &
            59.2 $\pm$ 0.4 &
            24.7$^*$ $\pm$ 2.4$\;\,$ &
            47.9$^*$ $\pm$ 1.2$\;\,$ &
            46.4$^*$ $\pm$ 0.4$\;\,$ &
            58.2$^*$ $\pm$ 0.1$\;\,$ &
            58.5 $\pm$ 0.4 \\
        \coloreddot{natural}~DTD & 
            70.8$^*$ $\pm$ 0.5$\;\,$ &
            54.3$^*$ $\pm$ 2.1$\;\,$ &
            72.8 $\pm$ 0.7 &
            60.5$^*$ $\pm$ 1.8$\;\,$ &
            68.1$^*$ $\pm$ 0.6$\;\,$ &
            68.4$^*$ $\pm$ 0.2$\;\,$ &
            73.0 $\pm$ 0.7 &
            72.6 $\pm$ 0.3 \\
        \coloreddot{natural}~Flowers102 & 
            99.3 $\pm$ 0.0 &
            76.7$^*$ $\pm$ 2.0$\;\,$ &
            99.1 $\pm$ 0.2 &
            86.5$^*$ $\pm$ 1.9$\;\,$ &
            95.3$^*$ $\pm$ 0.2$\;\,$ &
            96.7$^*$ $\pm$ 0.3$\;\,$ &
            99.2 $\pm$ 0.0 &
            99.2 $\pm$ 0.1 \\
        \coloreddot{natural}~Pets & 
            91.6 $\pm$ 0.3 &
            37.0$^*$ $\pm$ 0.9$\;\,$ &
            91.6 $\pm$ 0.3 &
            68.0$^*$ $\pm$ 1.5$\;\,$ &
            84.3$^*$ $\pm$ 0.8$\;\,$ &
            85.1$^*$ $\pm$ 1.9$\;\,$ &
            91.9 $\pm$ 0.5 &
            91.8 $\pm$ 0.4 \\
        \coloreddot{natural}~Sun397 & 
            47.1 $\pm$ 0.2 &
            \:\:1.1$^*$ $\pm$ 0.3$\;\,$ &
            47.2 $\pm$ 0.0 &
            \:\:1.0$^*$ $\pm$ 0.0$\;\,$ &
            36.3$^*$ $\pm$ 0.2$\;\,$ &
            38.3$^*$ $\pm$ 1.4$\;\,$ &
            47.7 $\pm$ 0.2 &
            47.4 $\pm$ 1.1 \\
        \coloreddot{natural}~SVHN & 
            35.1$^*$ $\pm$ 0.7$\;\,$ &
            78.9$^*$ $\pm$ 0.8$\;\,$ &
            74.3$^*$ $\pm$ 8.8$\;\,$ &
            82.6$^*$ $\pm$ 2.5$\;\,$ &
            88.6 $\pm$ 0.4 &
            85.8 $\pm$ 1.7 &
            82.7$^*$ $\pm$ 3.1$\;\,$ &
            85.2 $\pm$ 0.8 \\

        \midrule

        \coloreddot{specialized}~Camelyon &
            81.0$^*$ $\pm$ 0.3$\;\,$ &
            76.9$^*$ $\pm$ 0.7$\;\,$ &
            85.1 $\pm$ 1.1 &
            81.8$^*$ $\pm$ 1.4$\;\,$ &
            85.7 $\pm$ 0.8 &
            85.2 $\pm$ 0.5 &
            82.8$^*$ $\pm$ 1.2$\;\,$ &
            84.9 $\pm$ 0.9 \\
        \coloreddot{specialized}~EuroSAT &
            90.7$^*$ $\pm$ 0.5$\;\,$ &
            88.2$^*$ $\pm$ 2.2$\;\,$ &
            94.8 $\pm$ 0.5 &
            92.2$^*$ $\pm$ 1.0$\;\,$ &
            94.8 $\pm$ 0.3 &
            93.5 $\pm$ 0.6 &
            92.6$^*$ $\pm$ 0.8$\;\,$ &
            93.0 $\pm$ 0.3 \\
        \coloreddot{specialized}~Resisc45 & 
            72.9$^*$ $\pm$ 0.8$\;\,$ &
            66.5$^*$ $\pm$ 1.3$\;\,$ &
            83.5 $\pm$ 0.3 &
            74.4$^*$ $\pm$ 0.8$\;\,$ &
            82.5 $\pm$ 1.0 &
            79.6$^*$ $\pm$ 0.8$\;\,$ &
            80.0$^*$ $\pm$ 1.5$\;\,$ &
            82.0 $\pm$ 0.6 \\
        \coloreddot{specialized}~Retinopathy & 
            73.6$^*$ $\pm$ 0.0$\;\,$ &
            73.6$^*$ $\pm$ 0.0$\;\,$ &
            73.6$^*$ $\pm$ 0.0$\;\,$ &
            73.6$^*$ $\pm$ 0.0$\;\,$ &
            73.5$^*$ $\pm$ 0.1$\;\,$ &
            75.3 $\pm$ 1.2 &
            73.6$^*$ $\pm$ 1.1$\;\,$ &
            73.9 $\pm$ 0.4 \\

        \midrule

        \coloreddot{structured}~Clevr-Count &
            37.0$^*$ $\pm$ 0.9$\;\,$ &
            60.7$^*$ $\pm$ 6.9$\;\,$ &
            56.0$^*$ $\pm$ 2.9$\;\,$ &
            69.5 $\pm$ 4.0 &
            56.3$^*$ $\pm$ 1.9$\;\,$ &
            59.9$^*$ $\pm$ 2.6$\;\,$ &
            56.7$^*$ $\pm$ 1.2$\;\,$ &
            64.9 $\pm$ 2.2 \\
        \coloreddot{structured}~Clevr-Dist &
            34.4$^*$ $\pm$ 1.2$\;\,$ &
            56.7$^*$ $\pm$ 3.6$\;\,$ &
            52.5$^*$ $\pm$ 3.0$\;\,$ &
            60.3 $\pm$ 0.6 &
            59.0$^*$ $\pm$ 1.5$\;\,$ &
            57.3$^*$ $\pm$ 1.0$\;\,$ &
            57.5$^*$ $\pm$ 0.9$\;\,$ &
            60.3 $\pm$ 1.0 \\
        \coloreddot{structured}~DMLab & 
            37.7$^*$ $\pm$ 0.5$\;\,$ &
            37.2$^*$ $\pm$ 1.3$\;\,$ &
            47.0 $\pm$ 1.4 &
            44.5$^*$ $\pm$ 0.3$\;\,$ &
            50.0 $\pm$ 1.6 &
            44.9 $\pm$ 1.2 &
            42.9$^*$ $\pm$ 0.8$\;\,$ &
            45.2 $\pm$ 1.1 \\
        \coloreddot{structured}~dSpr-Loc & 
            12.8$^*$ $\pm$ 5.0$\;\,$ &
            65.0 $\pm$ 5.3 &
            40.4$^*$ $\pm$ 8.3$\;\,$ &
            69.3 $\pm$ 8.2 &
            76.5 $\pm$ 1.8 &
            73.6 $\pm$ 2.0 &
            62.8$^*$ $\pm$ 3.1$\;\,$ &
            64.1 $\pm$ 2.4 \\
        \coloreddot{structured}~dSpr-Ori & 
            28.7$^*$ $\pm$ 1.1$\;\,$ &
            31.2$^*$ $\pm$ 9.0$\;\,$ &
            45.0 $\pm$ 0.8 &
            49.5 $\pm$ 0.8 &
            49.0 $\pm$ 1.2 &
            48.7 $\pm$ 0.2 &
            41.2$^*$ $\pm$ 0.5$\;\,$ &
            44.8 $\pm$ 3.0 \\
        \coloreddot{structured}~KITTI-Dist & 
            59.3$^*$ $\pm$ 1.1$\;\,$ &
            68.2$^*$ $\pm$ 2.2$\;\,$ &
            82.4 $\pm$ 0.5 &
            76.5 $\pm$ 2.9 &
            82.7 $\pm$ 1.0 &
            82.0 $\pm$ 0.6 &
            79.5 $\pm$ 0.5 &
            78.0 $\pm$ 2.9 \\
        \coloreddot{structured}~sNORB-Azim & 
            14.1$^*$ $\pm$ 0.4$\;\,$ &
            11.7$^*$ $\pm$ 1.3$\;\,$ &
            16.5$^*$ $\pm$ 0.9$\;\,$ &
            16.7 $\pm$ 1.7 &
            18.7 $\pm$ 0.7 &
            17.7 $\pm$ 0.7 &
            14.9$^*$ $\pm$ 1.9$\;\,$ &
            16.9 $\pm$ 1.0 \\
        \coloreddot{structured}~sNORB-Elev & 
            23.5$^*$ $\pm$ 0.1$\;\,$ &
            20.7$^*$ $\pm$ 1.4$\;\,$ &
            26.9$^*$ $\pm$ 0.1$\;\,$ &
            33.8 $\pm$ 1.8 &
            31.1$^*$ $\pm$ 1.4$\;\,$ &
            29.3$^*$ $\pm$ 1.5$\;\,$ &
            29.1$^*$ $\pm$ 1.0$\;\,$ &
            32.1 $\pm$ 1.6 \\
    \bottomrule
    \end{tabular}}
    \caption{\textbf{Detailed VTAB-1k benchmark results using a ViT-S/14}. Average and standard deviation accuracies across three seeds for seven adapters across all 19 VTAB-1k datasets. Results marked with $^*$ indicate accuracies significantly lower ($p=0.05$) than the accuracy of PVeRA.}
    \label{tab:detailed_results_small}
\end{table*}

\begin{table*}[]
    \centering
    \resizebox{\textwidth}{!}{
    \begin{tabular}{l c c c c c c c c}
    \toprule
         &  Linear & Bottleneck & IA$^3$ & AdaptFormer & DoRA & LoRA & VeRA & PVeRA \\
    \midrule
        \coloreddot{natural}~Caltech101 &
            84.9$^*$ $\pm$ 0.5$\;\,$ &
            88.2 $\pm$ 0.5 & 
            85.0$^*$ $\pm$ 0.4$\;\,$ &
            86.4$^*$ $\pm$ 1.3$\;\,$ &
            88.1 $\pm$ 0.8 &
            89.0 $\pm$ 1.2 &
            89.9 $\pm$ 0.9 &
            88.5 $\pm$ 0.8 \\
        \coloreddot{natural}~CIFAR-100 &
            70.3$^*$ $\pm$ 0.1$\;\,$ &
            50.9$^*$ $\pm$ 35.3$\;\,$ &
            77.1 $\pm$ 0.1 &
            39.3$^*$ $\pm$ 25.5$\;\,$ &
            76.0$^*$ $\pm$ 0.9$\;\,$ &
            74.0 $\pm$ 0.5 &
            66.4$^*$ $\pm$ 1.3$\;\,$ &
            75.6 $\pm$ 1.4 \\
        \coloreddot{natural}~DTD & 
            74.0$^*$ $\pm$ 0.2$\;\,$ &
            77.6 $\pm$ 0.4 &
            76.6 $\pm$ 0.3 &
            74.8$^*$ $\pm$ 0.4$\;\,$ &
            79.0 $\pm$ 0.1 &
            78.0 $\pm$ 0.2 &
            75.4$^*$ $\pm$ 0.5$\;\,$ &
            77.1 $\pm$ 1.1 \\
        \coloreddot{natural}~Flowers102 &
            99.1$^*$ $\pm$ 0.2$\;\,$ &
            99.7 $\pm$ 0.0 &
            99.7 $\pm$ 0.0 &
            66.2$^*$ $\pm$ 46.3$\;\,$ &
            99.7 $\pm$ 0.0 &
            99.7 $\pm$ 0.0 &
            99.7 $\pm$ 0.0 &
            99.7 $\pm$ 0.0 \\
        \coloreddot{natural}~Pets &
            93.4 $\pm$ 0.9 &
            94.4 $\pm$ 0.2 &
            94.2 $\pm$ 1.1 &
            86.7$^*$ $\pm$ 2.0$\;\,$ &
            93.6 $\pm$ 1.2 &
            93.3 $\pm$ 0.6 &
            92.6$^*$ $\pm$ 0.8$\;\,$ &
            93.3 $\pm$ 1.2 \\
        \coloreddot{natural}~Sun397 &
            55.7$^*$ $\pm$ 0.4$\;\,$ &
            56.5$^*$ $\pm$ 0.4$\;\,$ &
            57.7$^*$ $\pm$ 0.1$\;\,$ &
            3.4$^*$ $\pm$ 0.5$\;\,$ &
            57.1$^*$ $\pm$ 0.2$\;\,$ &
            55.0$^*$ $\pm$ 0.4$\;\,$ &
            54.0$^*$ $\pm$ 1.2$\;\,$ &
            58.1 $\pm$ 0.1 \\
        \coloreddot{natural}~SVHN &
            32.9$^*$ $\pm$ 0.7$\;\,$ &
            67.4$^*$ $\pm$ 33.8$\;\,$ &
            44.9$^*$ $\pm$ 1.7$\;\,$ &
            30.1$^*$ $\pm$ 9.5$\;\,$ &
            81.1 $\pm$ 3.8 &
            92.4$^*$ $\pm$ 0.7$\;\,$ &
            89.2 $\pm$ 0.6 &
            88.9 $\pm$ 1.1 \\

        \midrule

        \coloreddot{specialized}~Camelyon &
            82.7$^*$ $\pm$ 0.5$\;\,$ &
            86.5 $\pm$ 0.4 &
            81.5$^*$ $\pm$ 0.4$\;\,$ &
            85.8$^*$ $\pm$ 1.8$\;\,$ &
            84.9 $\pm$ 0.6 &
            87.4$^*$ $\pm$ 1.3$\;\,$ &
            85.7$^*$ $\pm$ 1.2$\;\,$ &
            86.6 $\pm$ 1.0 \\
        \coloreddot{specialized}~EuroSAT &
            92.0$^*$ $\pm$ 0.1$\;\,$ &
            94.7$^*$ $\pm$ 0.3$\;\,$ &
            94.7$^*$ $\pm$ 0.2$\;\,$ &
            95.1 $\pm$ 0.3 &
            95.9$^*$ $\pm$ 0.3$\;\,$ &
            95.0 $\pm$ 0.1 &
            95.9 $\pm$ 0.2 &
            95.4 $\pm$ 0.2 \\
        \coloreddot{specialized}~Resisc45 &
            80.4$^*$ $\pm$ 0.8$\;\,$ &
            88.9$^*$ $\pm$ 1.7$\;\,$ &
            87.7$^*$ $\pm$ 0.3$\;\,$ &
            81.1$^*$ $\pm$ 4.9$\;\,$ &
            90.0 $\pm$ 0.4 &
            91.3 $\pm$ 0.3 &
            88.3$^*$ $\pm$ 0.7$\;\,$ &
            89.8 $\pm$ 0.2 \\
        \coloreddot{specialized}~Retinopathy &
            73.6 $\pm$ 0.0 &
            73.6 $\pm$ 0.0 &
            73.6 $\pm$ 0.0 &
            73.6 $\pm$ 0.0 &
            75.0 $\pm$ 1.0 &
            73.6 $\pm$ 0.0 &
            73.6 $\pm$ 0.0 &
            73.6 $\pm$ 0.0 \\

        \midrule

        \coloreddot{structured}~Clevr-Count &
            43.4$^*$ $\pm$ 1.5$\;\,$ &
            72.5$^*$ $\pm$ 6.6$\;\,$ &
            55.2$^*$ $\pm$ 1.5$\;\,$ &
            39.6$^*$ $\pm$ 13.1$\;\,$ &
            85.8$^*$ $\pm$ 4.5$\;\,$ &
            66.5 $\pm$ 3.5 &
            72.1$^*$ $\pm$ 6.0$\;\,$ &
            82.9 $\pm$ 4.2 \\
        \coloreddot{structured}~Clevr-Dist &
            31.9$^*$ $\pm$ 0.4$\;\,$ &
            60.4$^*$ $\pm$ 1.9$\;\,$ &
            52.3$^*$ $\pm$ 1.2$\;\,$ &
            41.4$^*$ $\pm$ 5.6$\;\,$ &
            57.5$^*$ $\pm$ 0.6$\;\,$ &
            58.9$^*$ $\pm$ 1.5$\;\,$ &
            60.3$^*$ $\pm$ 0.7$\;\,$ &
            62.7 $\pm$ 0.9 \\
        \coloreddot{structured}~DMLab &
            41.4$^*$ $\pm$ 0.7$\;\,$ &
            53.0 $\pm$ 2.5 & 
            46.6$^*$ $\pm$ 0.6$\;\,$ & 
            33.7$^*$ $\pm$ 2.5$\;\,$ &
            53.6 $\pm$ 0.5 & 
            56.5 $\pm$ 0.8 &
            53.2 $\pm$ 0.8 &
            52.5 $\pm$ 0.4 \\
        \coloreddot{structured}~dSpr-Loc &
            9.6$^*$ $\pm$ 0.6$\;\,$ &
            79.9 $\pm$ 6.7 &
            31.7$^*$ $\pm$ 1.2$\;\,$ &
            17.0$^*$ $\pm$ 5.2$\;\,$ &
            75.5$^*$ $\pm$ 3.0$\;\,$ &
            43.0 $\pm$ 7.2 &
            75.5 $\pm$ 4.0 &
            70.6 $\pm$ 7.0 \\
        \coloreddot{structured}~dSpr-Ori &
            25.5$^*$ $\pm$ 2.9$\;\,$ &
            53.1 $\pm$ 1.3 &
            47.6$^*$ $\pm$ 3.0$\;\,$ &
            18.3$^*$ $\pm$ 1.9$\;\,$ &
            55.8 $\pm$ 1.3 &
            53.4 $\pm$ 1.3 &
            52.3$^*$ $\pm$ 1.2$\;\,$ &
            53.0 $\pm$ 2.1 \\
        \coloreddot{structured}~KITTI-Dist &
            48.7$^*$ $\pm$ 3.4$\;\,$ &
            82.5$^*$ $\pm$ 1.7$\;\,$ &
            67.7$^*$ $\pm$ 1.0$\;\,$ &
            70.8$^*$ $\pm$ 2.9$\;\,$ &
            83.8 $\pm$ 1.2 &
            83.7 $\pm$ 1.5 &
            82.3$^*$ $\pm$ 1.1$\;\,$ &
            84.3 $\pm$ 1.2 \\
        \coloreddot{structured}~sNORB-Azim &
            12.6$^*$ $\pm$ 0.1$\;\,$ &
            21.6$^*$ $\pm$ 3.8$\;\,$ &
            12.6$^*$ $\pm$ 0.6$\;\,$ &
            7.7$^*$ $\pm$ 0.6$\;\,$ &
            21.4 $\pm$ 0.3 &
            26.2$^*$ $\pm$ 0.6$\;\,$ &
            24.0 $\pm$ 0.7 &
            24.2 $\pm$ 0.4 \\
        \coloreddot{structured}~sNORB-Elev &
            23.6$^*$ $\pm$ 0.9$\;\,$ &
            32.3$^*$ $\pm$ 2.6$\;\,$ &
            27.6$^*$ $\pm$ 0.0$\;\,$ &
            23.1$^*$ $\pm$ 1.7$\;\,$ &
            34.4$^*$ $\pm$ 0.2$\;\,$ &
            35.8$^*$ $\pm$ 1.6$\;\,$ &
            36.4 $\pm$ 1.7 &
            36.6 $\pm$ 2.2 \\
    \bottomrule
    \end{tabular}}
    \caption{
    \textbf{Detailed VTAB-1k benchmark results using a ViT-L/14}. Average and standard deviation accuracies across three seeds for seven adapters across all 19 VTAB-1k datasets. Results marked with $^*$ indicate accuracies significantly lower ($p=0.05$) than the accuracy of PVeRA.}
    \label{tab:detailed_results_large}
\end{table*}

\end{document}